%% file: iclr2024_conference.tex
\documentclass{article} 
\usepackage{iclr2024_conference,times}

\input{math_commands.tex}

\usepackage[utf8]{inputenc} 
\usepackage[T1]{fontenc}    
\usepackage{hyperref}       
\usepackage{url}            
\usepackage{booktabs}       
\usepackage{amsfonts}       
\usepackage{nicefrac}       
\usepackage{microtype}      
\usepackage{algorithm}
\usepackage{algcompatible}
\usepackage{setspace}
\usepackage{listings}
\usepackage{xcolor}
\usepackage{amssymb}
\usepackage{float}
\usepackage{graphicx}
\usepackage{amsmath}
\usepackage{wrapfig}

\usepackage{subfigure}
\usepackage{parskip}
\usepackage{makecell}
\usepackage{marvosym}
\usepackage{multirow}
\usepackage{pifont}
\usepackage{enumitem}
\usepackage{svg} 

\usepackage{array,multicol,tabularx}

\title{Efficient ConvBN Blocks for Transfer \\ Learning and Beyond}

\makeatletter
\newcommand{\thanksone}[1]{\textsuperscript{$*$}\protected@xdef\@thanks{\@thanks\protect\footnotetext{$*$:#1}{}}}
\newcommand{\thankstwo}[1]
{\textsuperscript{\Letter{}}\protected@xdef\@thanks{\@thanks\protect\footnotetext{\Letter{}:#1}{}}}
\makeatother

\newcommand\pd[2]{\frac{\partial {#1}}{\partial {#2}}}
\newcommand{\tabincell}[2]{\begin{tabular}{@{}#1@{}}#2\end{tabular}}

\author{Kaichao You$^\dagger$\thanksone{\; This work is conducted during Kaichao You’s internship at Apple.}, Guo Qin$^\dagger$, Anchang Bao$^\dagger$ Meng Cao$^\S$, Ping Huang$^\S$, Jiulong Shan$^\S$, \\ \textbf{Mingsheng Long}$^\dagger$\thankstwo{\; Mingsheng Long  is the corresponding author.} \\
$^\dagger$ School of Software, BNRist, Tsinghua University, China \; \; $^\S$ Apple\\
\texttt{\{ykc20,bac20,qing20\}@mails.tsinghua.edu.cn} \\
\texttt{\{mengcao,Huang\_ping,jlshan\}@apple.com} \\
\texttt{mingsheng@tsinghua.edu.cn}
}

\iclrfinalcopy 
\begin{document}

\maketitle

\begin{abstract}
  Convolution-BatchNorm (ConvBN) blocks are integral components in various computer vision tasks and other domains. A ConvBN block can operate in three modes: Train, Eval, and Deploy. While the Train mode is indispensable for training models from scratch, the Eval mode is suitable for transfer learning and beyond, and the Deploy mode is designed for the deployment of models. This paper focuses on the trade-off between stability and efficiency in ConvBN blocks: Deploy mode is efficient but suffers from training instability; Eval mode is widely used in transfer learning but lacks efficiency. To solve the dilemma, we theoretically reveal the reason behind the diminished training stability observed in the Deploy mode. Subsequently, we propose a novel Tune mode to bridge the gap between Eval mode and Deploy mode. The proposed Tune mode is as stable as Eval mode for transfer learning, and its computational efficiency closely matches that of the Deploy mode. Through extensive experiments in object detection, classification, and adversarial example generation across $5$ datasets and $12$ model architectures, we demonstrate that the proposed Tune mode retains the performance while significantly reducing GPU memory footprint and training time, thereby contributing efficient ConvBN blocks for transfer learning and beyond. Our method has been integrated into both PyTorch (general machine learning framework) and MMCV/MMEngine (computer vision framework). Practitioners just need one line of code to enjoy our efficient ConvBN blocks thanks to PyTorch's builtin machine learning compilers.

\end{abstract}

\section{Introduction}


Feature normalization~\citep{huang_normalization_2023} is a critical component in deep convolutional neural networks to facilitate the training process by promoting stability, mitigating internal covariate shift, and enhancing network performance. BatchNorm~\citep{ioffe_batch_2015} is a popular and widely adopted normalization module in computer vision. A convolutional layer~\citep{lecun_gradient-based_1998} together with a consecutive BatchNorm layer is often called a ConvBN block, which operates in three modes:
\begin{itemize}
    \item Train mode. Mini-batch statistics (mean and standard deviation $\mu, \sigma$) are computed for feature normalization, and running statistics ($\hat{\mu}, \hat{\sigma}$) are tracked by exponential moving averages for testing individual examples when mini-batch statistics are unavailable.
    \item Eval mode. Running statistics are directly used for feature normalization without update, which is more efficient than Train mode, but requires tracked statistics to remain stable in training. It can also be used to validate models during development.
    \item Deploy mode. When the model does not require further training, computation in Eval mode can be accelerated~\citep{markus_fusing_nodate} by fusing convolution, normalization, and affine transformations into a single convolutional operator with transformed parameters. This is called Deploy mode, which produces the same output as Eval mode with better efficiency. In Deploy mode, parameters for the convolution are computed once-for-all, removing batch normalization for faster inference during deployment.
\end{itemize}
The three modes of ConvBN blocks present a trade-off between computational efficiency and training stability, as shown in Table~\ref{tab:trade-off}. Train mode is applicable for both train from scratch and transfer learning, while Deploy mode optimizes computational efficiency. Consequently, these modes traditionally align with three stages in deep models' lifecycle: Train mode for training, Eval mode for validation, and Deploy mode for deployment.
\begin{table}[htbp]
    \centering
    \vspace{-10pt}
    \caption{Trade-off among modes of ConvBN blocks.}
      \begin{tabular}{ccccc}
      \toprule
         Mode   & Train & Eval & Tune (proposed) & Deploy \\
            \midrule
      \textcolor{black!50}{Train From Scratch} &   \textcolor{black!50}{\ding{51}}    &   \textcolor{black!50}{\ding{55}}    &    \textcolor{black!50}{\ding{55}}   & \textcolor{black!50}{\ding{55}} \\
      \midrule
      Transfer Learning &   \ding{51}  &   \ding{51}    &      \ding{51} & \ding{55} \\
      \midrule
      Training Efficiency &    {\LARGE{$\star$}}   &   {\LARGE{$\star$}} {\LARGE{$\star$}}    &   {\LARGE{$\star$}} {\LARGE{$\star$}} {\LARGE{$\star$}}   & {\LARGE{$\star$}} {\LARGE{$\star$}} 
 {\LARGE{$\star$}} \\
      \bottomrule
      \end{tabular}%
    \label{tab:trade-off}%
\end{table}%

\begin{figure}[h]
    \centering
    \includegraphics[width=.8\textwidth]{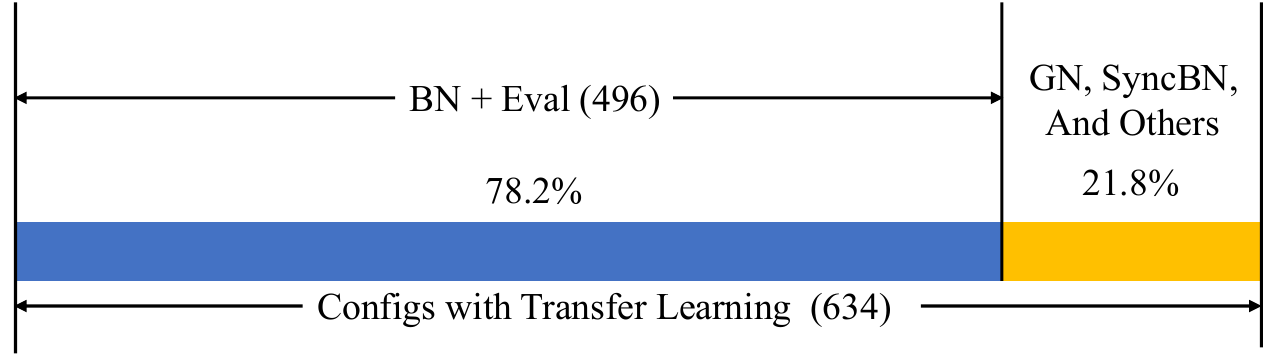}
    \vspace{-10pt}
    \caption{Usage of normalization layers in all the 634 object detectors with pre-trained backbones in the MMDetection framework~\citep{chen_mmdetection:_2019}. GN denotes GroupNorm, SyncBN represents synchronized BatchNorm across multiple GPUs and Eval indicates training ConvBN blocks in Eval mode. A majority of detectors (over 78\%) are trained with ConvBN blocks in Eval mode.}
    \label{fig:mmdet_statistics}
\end{figure}

With the rise of transfer learning~\citep{jiang2022transferability}, practitioners usually start with a pre-trained model, and instability of training from scratch is less of a concern. For instance, an object detector typically has one pre-trained backbone to extract features, and a head trained from scratch to predict bounding boxes and categories. Therefore, practitioners have started to explore Eval mode for transfer learning, which is more efficient than Train mode. Figure~\ref{fig:mmdet_statistics} presents the distribution of the normalization layers used in MMDetection~\citep{chen_mmdetection:_2019}, a popular object detection framework. In the context of transfer learning, a majority of detectors (496 out of 634) are trained with ConvBN blocks in Eval mode. Interestingly, our experiments suggest that Eval mode not only improves computational efficiency but also enhances the final performance over Train mode in certain transfer learning scenarios. For example, Appendix~\ref{appendix:compare_train_eval} shows training Faster-RCNN~\citep{ren_faster_2015} on COCO~\citep{lin_microsoft_2014} with Eval mode achieves significantly better mAP than Train mode, with either pre-trained ResNet101 backbone or pre-trained HRNet backbone.

Since transfer learning of ConvBN blocks in Eval mode has been a common practice, and forward calculation results between Deploy mode and Eval mode are equivalent, it is natural to ask if we can use Deploy mode for more efficient training. Unfortunately, Section~\ref{sec:compare_eval_and_deploy} shows direct training in Deploy mode can lead to instability, as it is not designed for training.

In quest of efficient ConvBN blocks for transfer learning with pre-trained models, we theoretically uncover the underlying causes of training instability in Deploy mode and subsequently propose a novel Tune mode. It bridges the gap between Eval mode and Deploy mode, preserving functional equivalence with Eval mode in both forward and backward propagation while approaching the computational efficiency of Deploy mode. Our extensive experiments across transfer learning tasks (object detection and classification) and beyond (adversarial example generation) confirm the reduction in memory footprint and wall-clock training time without sacrificing performance.

Our contributions are summarized as follows:

\begin{itemize}
    \setlength\itemsep{-3pt}
    \item We theoretically analyze why Deploy mode is unstable for training, and propose efficient ConvBN blocks with a Tune mode to take advantages from both Eval and Deploy modes.
    \item We present extensive experiments across $12$ models and $5$ datasets to confirm the gain of Tune mode in transfer learning and beyond.
    \item Our method has been quickly integrated into open-source framework libraries like PyTorch and MMCV/MMEngine because of its evident benefit, improving the efficiency of hundreds of models for everyone using these frameworks.
\end{itemize}

\section{Related Work}

\subsection{Normalization Layers}

Feature normalization has long been established in machine learning~\citep{bishop_pattern_2006}, \emph{e.g.}, Z-score normalization to standardize input features for smooth and isotropic optimization landscapes~\citep{boyd_convex_2004}. With the emergence of deep learning, normalization methods specifically tailored to intermediate activations, or feature maps, have been developed and gained traction.

Batch normalization (BN), proposed by \citet{ioffe_batch_2015}, demonstrated that normalizing intermediate layer activations could expedite training and mitigate the effects of internal covariate shift. Since then, various normalization techniques have been proposed to address specific needs, such as group normalization~\citep{wu_group_2018} for small batch sizes, and layer normalization~\citep{ba_layer_2016} typically employed in sequence models such as recurrent networks and Transformers. We direct interested readers to the survey by \citet{huang_normalization_2023} for an in-depth exploration of normalization layers.

Among various types of normalization, BatchNorm is a popular choice, partly due to its ability to be fused within convolution operations during deployment. Conversely, other normalization layers exhibit different behaviors compared to BatchNorm and often entail higher computational costs during deployment. The fusion allows ConvBN blocks to be efficiently deployed to massive edge and mobile devices where efficiency and power consumption control are critically important. This paper focuses on improving the efficiency of widely used ConvBN blocks for transfer learning and beyond.

\subsection{Variants of Batch Normalization}

While batch normalization successfully improves training stability and convergence, it presents several limitations. These limitations originate from the different behavior during training and validation (Train mode and Eval mode), which is referred to as train-inference mismatch~\citep{gupta_no-frills_2019} in the literature. \citet{ioffe_batch_2017} proposed batch renormalization to address the normalization issues with small batch sizes. \citet{wang_transferable_2019} introduced TransNorm to tackle the normalization problem when adapting a model to a new domain. Recently, researchers find that train-inference mismatch of BatchNorm plays an important role in test-time domain adaptation~\citep{wang_tent:_2021, wang_continual_2022}. In this paper, we focus on training ConvBN blocks in Eval mode, which is free of train-inference mismatch because its behavior is consistent in both training and inference.

Another challenge posed by BatchNorm is its memory intensiveness. While the computation of BatchNorm is relatively light compared with convolution, it occupies nearly the same memory as convolution because it records the feature map of convolutional output for back-propagation. To address this issue, \citet{bulo_-place_2018} proposed to replace the activation function (ReLU~\citep{nair_rectified_2010}) following BatchNorm with an invertible activation (such as Leaky ReLU~\citep{maas_rectifier_2013}), thereby eliminating the need to store the output of convolution for backpropagation. However, this approach imposes an additional computational burden on backpropagation, as the input of activations must be recomputed by inverting the activation function. Their reduced memory footprint comes with the price of increased running time. In contrast, our proposed Tune mode effectively reduces both computation time and memory footprint for efficient transfer learning without any modification of network activations or any other architecture. 

\subsection{Transfer Learning}

Training deep neural networks used to be difficult and time-consuming. Fortunately, with the advent of advanced network architectures like skip connections~\citep{he_deep_2016}, and the availability of foundation models~\citep{bommasani_opportunities_2022}, practitioners can now start with pre-trained models and fine-tune them for various applications. Pre-trained models offer general representations~\citep{donahue_decaf:_2014} that can accelerate the convergence of fine-tuning in downstream tasks. Consequently, the rule of thumb in computer vision tasks is to start with models pre-trained on large-scale datasets like ImageNet~\citep{deng_imagenet:_2009}, Places~\citep{zhou_places:_2018}, or OpenImages~\citep{kuznetsova_open_2018}. This transfer learning paradigm alleviates the data collection burden required to build a deep model with satisfactory performance and can also expedite training, even if the downstream task has abundant data~\citep{mahajan_exploring_2018}.

Train mode is the only mode for training ConvBN blocks from scratch. However, it is possible to use Eval mode for transfer learning, as we can exploit pre-trained statistics without updating them. Moreover, Eval mode is more computationally efficient than Train mode. Consequently, researchers usually fine-tune pre-trained models in Eval mode~\citep{chen_mmdetection:_2019} (Figure~\ref{fig:mmdet_statistics}), which maintains performance while offering improved computational efficiency. In this paper, we propose a novel Tune mode that further reduces memory footprint and training time while maintaining functional equivalence with the Eval mode during both forward and backward propagation. The Tune mode is a drop-in replacement of Eval mode while making ConvBN blocks more efficient.

\subsection{Machine Learning Compilers}

PyTorch~\citep{paszke_pytorch:_2019}, a widely adopted deep learning framework, uses dynamic computation graphs that are built on-the-fly during computation. This imperative style of computation is user-friendly and leads to PyTorch's rapid rise in popularity. Nevertheless, the dynamic computation graphs complicate the speed optimization. Traditionally, operator analysis and fusion were only applied to models after training. The speed optimization usually involved a separate language or framework such as TensorRT~\citep{vanholder_efficient_2016} or other domain-specific languages, distinct from the Python language commonly employed for training. PyTorch has explored several ways, including symbolic tracing with \texttt{torch.fx}~\citep{reed_torch._2022} and just-in-time tracing with \texttt{torch.jit}, to introduce machine learning compilers into the framework, and all the efforts are consolidated into PyTorch 2.0~\citep{wu_pytorch_2023}. Leveraging PyTorch's pioneering compiler, our proposed Tune mode can automatically identify consecutive Convolution and BatchNorm layers without manual intervention.

\section{Method}

\subsection{Problem Setup}

In this paper, we study ConvBN blocks that are prevalent in various computer vision applications, especially in edge and mobile devices. A ConvBN block consists of two layers: (1) a convolutional layer with weight $\omega$ and bias $b$; (2) a BatchNorm layer with tracked mean $\hat{\mu}$ and standard deviation $\hat{\sigma}$, and weight $\gamma$ and bias $\beta$. We focus on the computation within ConvBN blocks, which is not affected by activation functions or skip connections after ConvBN blocks.

Given an input tensor $X$ with dimensions $[N, C_\text{in}, H_\text{in}, W_\text{in}]$, where $N$ represents the batch size, $C_\text{in}$ the number of input channels, and $H_\text{in}/W_\text{in}$ the spatial height/width of the input, a ConvBN block in Eval mode (the majority choice in transfer learning as shown in Figure~\ref{fig:mmdet_statistics}) operates as follows. First, the convolutional layer computes an intermediate output tensor $Y = \omega \circledast X + b$ (we use $\circledast$ to denote convolution), resulting in dimensions $[N, C_\text{out}, H_\text{out}, W_\text{out}]$. Subsequently, the BN layer normalizes and applies an affine transformation to the intermediate output, producing the final output tensor $Z=\gamma \frac{Y-\hat{\mu}}{\sqrt{\hat{\sigma}^2+\varepsilon}}+\beta$ with the same dimensions as $Y$.

Usually, the training loss consists of two parts: $J = J(Z)$ calculated on the network's output, and regularization loss $R$ calculated on the network's trainable parameters. Training is dominated by the gradient from $J(Z)$, especially at the beginning of training. The influence of $R$ is rather straightforward to analyze, since it directly and independently applies to each parameter. We omit the analysis for simplicity, as it does not change the main conclusion of this paper. Therefore, our primary focus lies in understanding the gradient with respect to the output loss function $J(Z)$ under different modes of ConvBN blocks. Note that $J(Z)$ can represent loss directly calculated on $Z$, as well as loss computed based on the output of subsequent layers operating on $Z$.

\subsection{Preliminary}

\subsubsection{Backward Propagation of Convolution}

To discuss the stability of training, we must examine the details of backward propagation to understand the behavior of the gradient for each parameter. For a convolution layer with forward computation $Y = \omega \circledast X + b$, if the gradient back-propagated to $Y$ is $\pd{J}{Y}$, then the gradients of each input of the convolution layer, as explained in~\citet{bouvrie_notes_2006}, are: $\pd{J}{\omega} = \pd{J}{Y} \odot X ; \pd{J}{X} = \omega_{\text{rot}} \circledast \pd{J}{Y} ; \pd{J}{b} = \pd{J}{Y}$. The $\odot$ represents cross-correlation, and $\omega_{\text{rot}}$ is the rotated version of $\omega$, both are used to compute the gradient of convolution~\citep{rabiner_theory_1975}. Note that these equations potentially contain broadcasting, a technique to allow element-wise arithmetic between two tensors with different shapes. Appendix~\ref{sec:broadcast} clarifies how broadcasting works in details.

\subsubsection{Associative Law for Convolution and Affine Transform}
Convolution can essentially be viewed as a patch-wise matrix-vector multiplication, with the matrix (kernel weight) having a shape of $[C_\text{out}, k^2 C_\text{in}]$, and the vector having a shape of $[k^2 C_\text{in}]$. If an affine transform is applied to the weight along the $C_\text{out}$ dimension, then the affine transform is associative with the convolution operator. Formally speaking, $\gamma \cdot (\omega \circledast X) = (\gamma \cdot \omega) \circledast X$, where $\gamma$ is a $C_\text{out}$-dimensional vector multiplied to each row of the weight $\omega$. This association law lays the foundation of analyses for the Deploy mode and our proposed Tune mode. The associative law also applies to transposed convolution~\citep{zeiler_deconvolutional_2010} and linear layers, therefore \emph{the proposed Tune mode also works for TransposedConv-BN and Linear-BN blocks}.

With the necessary background established, we directly present the forward, backward, and memory footprint details in Table~\ref{tab:conv_bn_details}. Further analyses will be provided in subsequent sections. 

\begin{table}[h]
    \centering
    \vspace{-10pt}
    \caption{Computation graph of ConvBN blocks in different modes. Shape annotations for each tensor are available in Appendix~\ref{appendix:code_detail_3_mode}. We introduce Tune mode to improve the efficiency of ConvBN blocks, alleviating the dilemma between training stability and computational efficiency.}
        \label{tab:conv_bn_details}
    
     \resizebox{\textwidth}{!}{
    \begin{tabular}{ccl}
        \toprule
            Mode & \tabincell{c}{Computation \\ Graph} & \makecell[c]{Backward \\ Propagation} \\
        \midrule
        Eval &\begin{minipage}{0.6\columnwidth}
		\centering
  {\includegraphics[width=\linewidth]{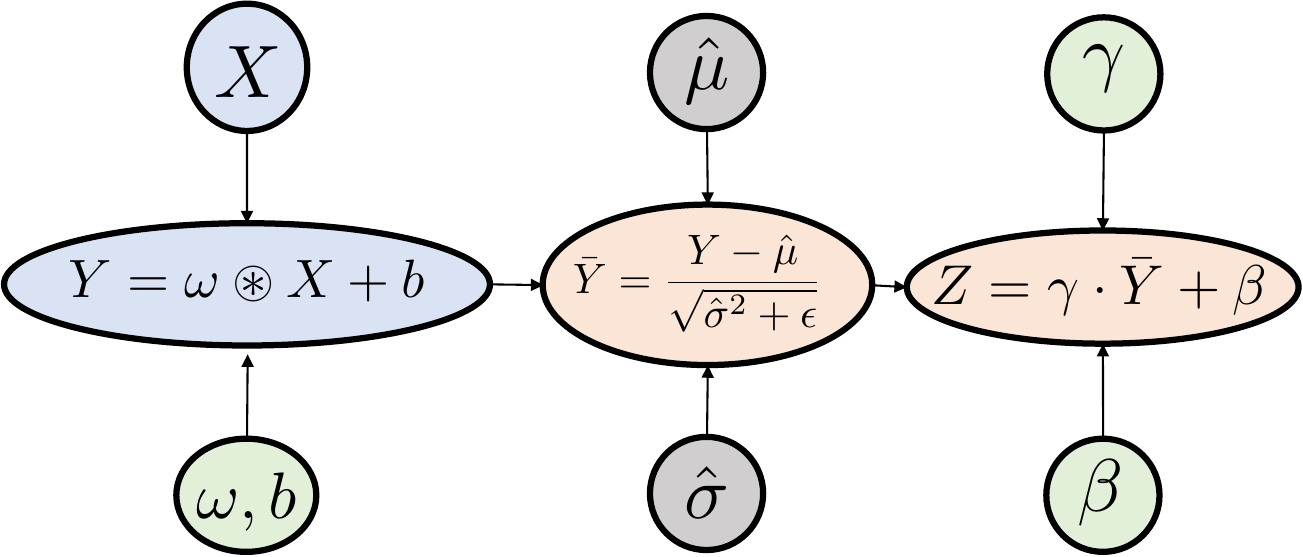}}
	\end{minipage}& $
\begin{small}\begin{aligned}
    &\frac{\partial J}{\partial X} = \frac{\gamma}{\sqrt{\hat{\sigma}^2 + \epsilon}} \omega_{\text{rot}} \circledast \frac{\partial J}{\partial Z} \\[2mm]
    &\frac{\partial J}{\partial \omega} = \frac{\gamma}{\sqrt{\hat{\sigma}^2 + \epsilon}} \pd{J}{Z} \odot X \\[2mm]
    &\pd{J}{b} = \frac{\gamma}{\sqrt{\hat{\sigma}^2 + \epsilon}} \pd{J}{Z} \\[2mm]
    &\pd{J}{\gamma} = \pd{J}{Z} \frac{Y - \hat{\mu}}{\sqrt{\hat{\sigma}^2 + \epsilon}} \\[2mm]
    &\pd{J}{\beta} = \pd{J}{Z}
\end{aligned}\end{small}
$
        \\
\midrule
            \tabincell{c}{Tune \\ (Proposed)} &
            \begin{minipage}{0.4\columnwidth}
		\centering
  {\includegraphics[width=\linewidth]{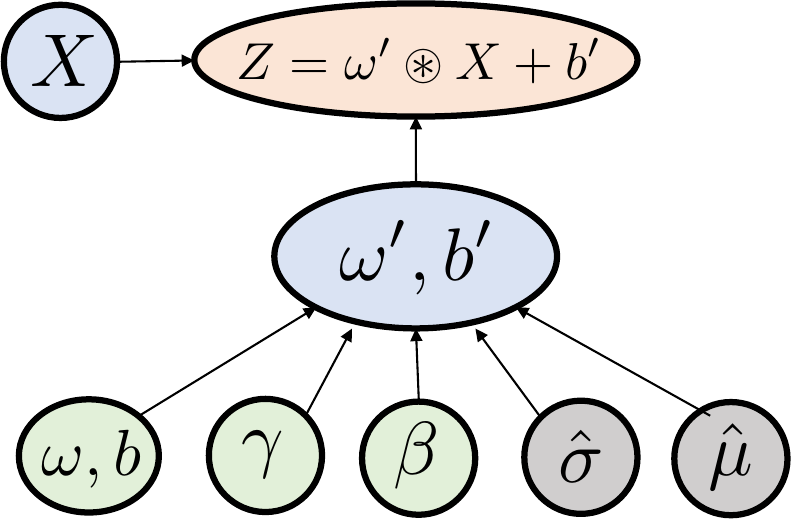}}
	\end{minipage} &$\begin{small}
\begin{aligned}
    &\pd{J}{X} = \omega_{\text{rot}}^\prime \circledast \pd{J}{Z} \\[2mm]
    &\pd{J}{\omega^\prime} = \pd{J}{Z} \odot X \\[2mm]
    &\pd{J}{\omega} = \frac{\gamma}{\sqrt{\hat{\sigma}^2 + \epsilon}} \pd{J}{\omega^\prime}\\[2mm]
    &\pd{J}{b} = \frac{\gamma}{\sqrt{\hat{\sigma}^2 + \epsilon}} \pd{J}{Z} \\[2mm]
    &\pd{J}{\gamma} = \pd{J}{\omega^\prime}\frac{\omega}{\sqrt{\hat{\sigma}^2 + \epsilon}} + \pd{J}{Z}\frac{b - \hat{\mu}}{\sqrt{\hat{\sigma}^2 + \epsilon}} \\[2mm]
    &\pd{J}{\beta} = \pd{J}{Z}
\end{aligned}\end{small}
$
\\
    \midrule
Deploy & \begin{minipage}{0.4\columnwidth}
		\centering
  {\includegraphics[width=\linewidth]{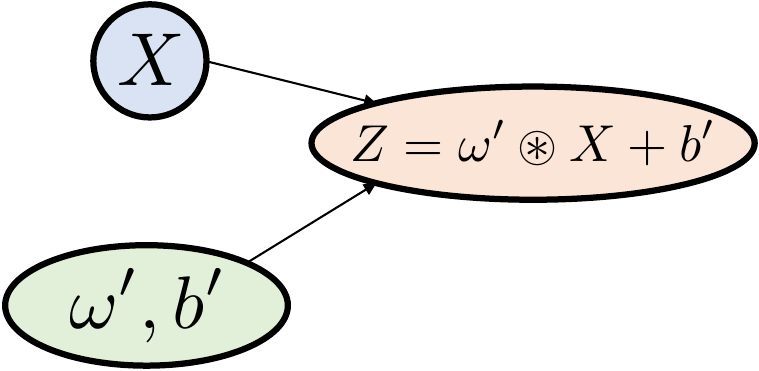}}
	\end{minipage} & $\begin{small}
\begin{aligned}
    &\pd{J}{X} = \omega_{\text{rot}}^\prime \circledast \pd{J}{Z} \\[2mm]
    &\pd{J}{\omega^\prime} = \pd{J}{Z} \odot X \\[2mm]
    &\pd{J}{b^\prime} = \pd{J}{Z}
\end{aligned}\end{small}
$ 
 \\
 \bottomrule \\
 \multicolumn{3}{c}{\begin{minipage}{\columnwidth}
		\centering
  {\includegraphics[width=\linewidth]{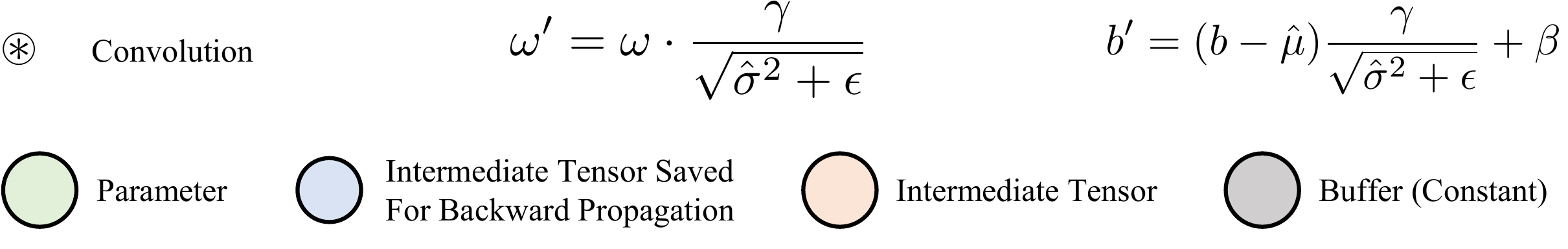}}
	\end{minipage}}
        
    \end{tabular}

}

\end{table}

\subsection{Analyzing Eval Mode and Deploy Mode}
\label{sec:compare_eval_and_deploy}

With the help of equations in Table~\ref{tab:conv_bn_details}, the comparison between Eval mode and Deploy mode on efficiency and training stability is as straightforward as follows.

\subsubsection{Forward Computation Efficiency} We first observe that Eval mode and Deploy mode have equivalent results in forward computation, and Deploy mode is more efficient. The equivalence can be proved by the definitions $\omega^{\prime}=\frac{\gamma}{\sqrt{\hat{\sigma}^2+\epsilon}} \cdot \omega$ and $b^{\prime}=\left(b-\hat{\mu}\right) \frac{\gamma}{\sqrt{\hat{\sigma}^2+\epsilon}}+\beta $, together with the associative law for convolution and affine transformations. However, Deploy mode pre-computes the weight $\omega^{\prime}$ and $b^{\prime}$, reducing the forward propagation to a single convolution calculation. Conversely, Eval mode requires a convolution, supplemented by a normalization and an affine transform on the convolutional output. This results in a slower forward propagation process for Eval mode. Moreover, Eval mode requires storing $X, Y$ for backward propagation, while Deploy mode only stores $X$. The memory footprint of Eval mode is nearly double of that in Deploy mode. Therefore, Deploy mode emerges as the more efficient of the two in terms of memory usage and computational time.

\subsubsection{Training Stability} 
\begin{figure}[h]
    \subfigure[]{\includegraphics[width=0.33\columnwidth]{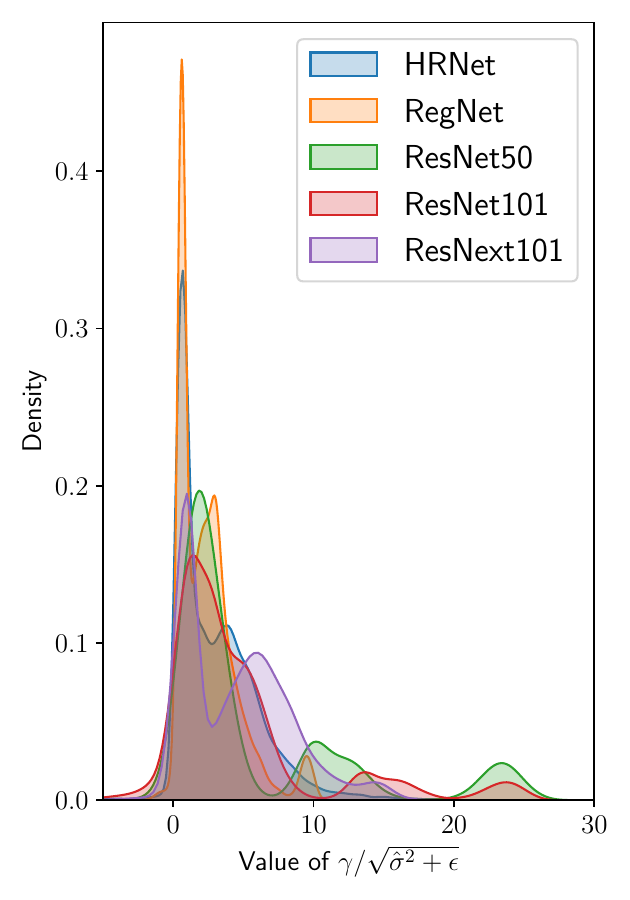}\label{fig:coeff_distribution}}
    \subfigure[]{\includegraphics[width=0.64\columnwidth]{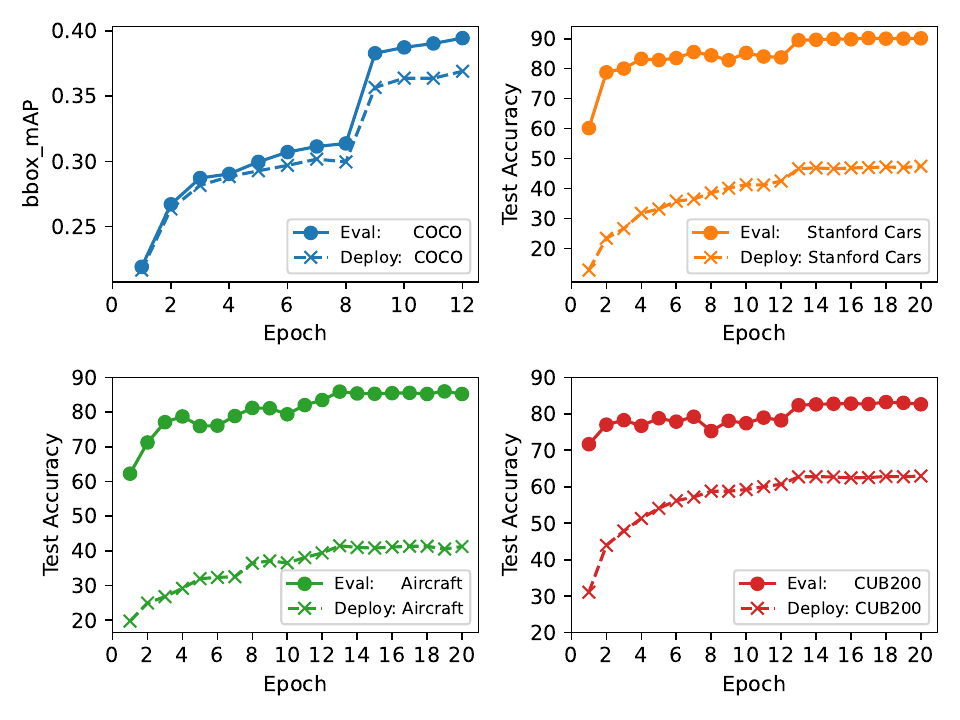}\label{fig:deploy_mode_failure}}
    \vspace{-5pt}
    \caption{(a): Distribution of scaling coefficients for weight $\left(\gamma/\sqrt{\hat{\sigma}^2 + \epsilon}\right)$ in different backbones. (b): Comparison between training with Eval mode and Deploy mode in both object detection and classification. Severe performance degradation is observed for training with Deploy mode.}
    \label{fig:coeff_distribution_and_deploy_mode_failure}
\end{figure}

Our analyses suggest that Deploy mode tends to exhibit less training stability than Eval mode. Focusing on the convolutional weight, which constitutes the primary parameters in ConvBN blocks, we observe from Table~\ref{tab:conv_bn_details} that the relationship of values and gradients between Deploy mode and Eval mode is $\omega^{\prime}=\frac{\gamma}{\sqrt{\hat{\sigma}^2+\epsilon}} \omega$ and $\frac{\partial J}{\partial \omega^{\prime}} = \frac{\sqrt{\hat{\sigma}^2+\epsilon}}{\gamma} \frac{\partial J}{\partial \omega}$. The scaling coefficients of the weight ($\frac{\gamma}{\sqrt{\hat{\sigma}^2+\epsilon}}$) are inverse of the scaling coefficients of the gradient ($\frac{\sqrt{\hat{\sigma}^2+\epsilon}}{\gamma}$). This can cause training instability in Deploy mode. For instance, if $\frac{\gamma}{\sqrt{\hat{\sigma}^2+\epsilon}}$ is small (say $0.1$), the weight reduces to one-tenth of its original value, while the gradient increases tenfold. This is a significant concern in real-world applications. As illustrated in Figure~\ref{fig:coeff_distribution}, these scaling coefficients range from as low as $0$ to as high as $30$, leading to unstable training. Figure~\ref{fig:deploy_mode_failure}  further substantiates this point through end-to-end experiments in both object detection and classification using Eval mode and Deploy mode. Training performance in Deploy mode is markedly inferior to that in Eval mode.

In conclusion, \emph{Deploy mode and Eval mode share the same forward calculation results, but present a dilemma in computation efficiency and training stability.}

\subsection{Tune Mode \emph{v.s.} Deploy Mode and Eval Mode}

Table~\ref{tab:conv_bn_details} describes the detailed computation of the proposed Tune mode specifically designed for efficient transfer learning. This mode leverages the associative law of convolution and affine transformation, optimizing both memory and computation. \emph{The main point is to calculate the transformed parameters dynamically, on-the-fly.} Next, we provide two critical analyses to show how the proposed Tune mode addresses the dilemma between training stability and computational efficiency, and how it bridges the gap between Eval mode and Deploy mode.

\subsubsection{Training Stability}

The associative law between convolution and affine transformation readily implies that the forward calculations between Eval mode and Tune mode are equivalent. The equivalence of backward calculations is less intuitive, particularly when considering the gradient of $\gamma$. To validate this, we employ an alternative approach: let $Z_1, Z_2$ represent the outputs of Eval mode and Tune mode, respectively. We define $Z_1 = Z_1 (\omega, b, \gamma, \beta), Z_2 = Z_2 (\omega, b, \gamma, \beta)$. Given that $Z_1 = Z_2$, and both are functions computed from the same set of parameters ($\omega, b, \gamma, \beta$), we can assert that their Jacobian matrices are the same: $\pd{Z_1}{[\omega, b, \gamma, \beta]} = \pd{Z_2}{[\omega, b, \gamma, \beta]}$. This immediately suggests that both modes share the same backward propagation dynamics. Consequently, we can conclude that \emph{Tune mode is as stable as Eval mode in transfer learning}. 

\subsubsection{Efficiency}

According to Table~\ref{tab:conv_bn_details}, Eval mode requires saving the input feature map $X$ and the convolutional output $Y$, with total memory footprint $X + Y$ for each ConvBN block. In contrast, Tune mode stores $X$ and the transformed weights $\omega^{\prime}$, with total memory footprint $X + \omega^{\prime}$ for each ConvBN block. Since feature maps $Y$ are usually larger than convolutional weights $\omega^{\prime}$, this difference signifies that Tune mode requires less memory for training. The same applies to the analysis of computation: computation in Eval mode consists of a convolution followed by an affine transformation on the \emph{convolutional feature map} $Y$; Tune mode computation consists of an affine transformation on the original \emph{convolutional weights} $\omega$ succeeded by a convolution with the transformed weights $\omega^\prime$. An affine transformation on convolutional weights executes faster than on feature maps. Therefore, Tune mode outperforms Eval mode both in memory usage and computation speed. Please refer to Appendix~\ref{appendix:theoretical} for formal analyses of efficiency using the $\mathcal{O}$ notation.

\begin{figure}[h]
    \includegraphics[width=\textwidth]{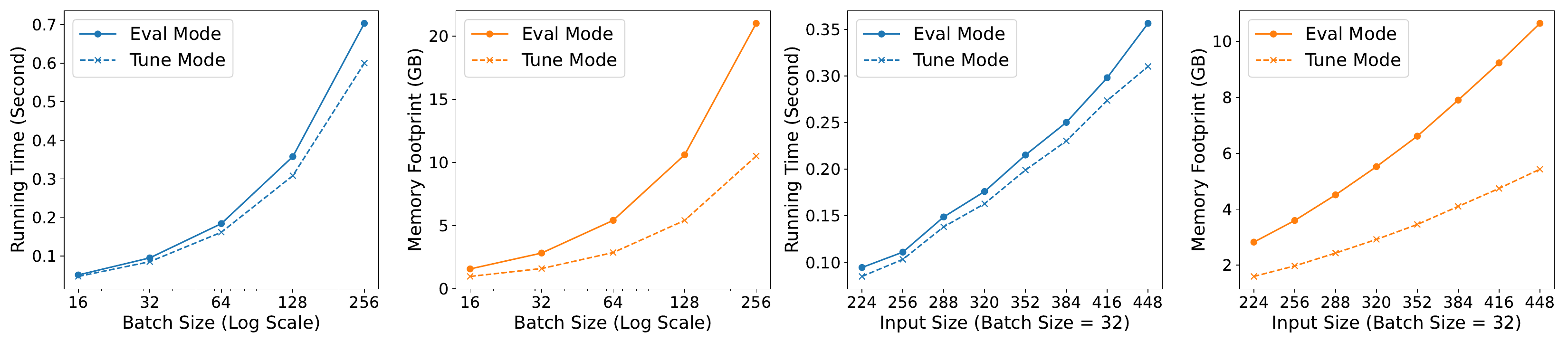}
    \vspace{-10pt}
    \caption{Memory footprint and running time comparison for Eval mode and Tune mode. The base setting is batchsize $=32$ and input dimension $=224 \times 224$, and we vary batchsize and input dimension to test the efficiency.}
    \label{fig:compute_efficiency}
\end{figure}

The above conclusion can be empirically validated using a standard ResNet-50~\citep{he_deep_2016} model with variable batch sizes and input sizes. The results, displayed in Figure~\ref{fig:compute_efficiency}, clearly indicate that \emph{Tune mode is more efficient than Eval mode} across all tested settings. The memory footprint of Tune mode consumed by pre-trained backbone in transfer learning can be reduced to one half of that in Eval mode, and the computation time is reduced by about $10\%$. The comparison between Tune and Deploy in efficiency can be found in Appendix~\ref{sec:tune_and_deploy_compare}, they are nearly the same in terms of efficiency, but Deploy mode is less stable and incurs worse accuracy than Tune mode.

The comparison among Eval/Tune/Deploy can be summarized as follows:

\begin{itemize}
    \setlength\itemsep{-3pt}
    \item $\text{Deploy mode} \approx \text{Tune mode} > \text{Eval mode}$, in terms of efficiency.
    \item $\text{Eval mode} = \text{Tune mode} > \text{Deploy mode}$, in terms of training stability.
\end{itemize}

Therefore, the proposed Tune mode successfully bridges the gap between Eval mode and Tune mode, improving the efficiency of ConvBN blocks with Eval mode while keeping the stability of training.

\section{Experiments}
\label{sec:experiments}

Our algorithm has been tested against $5$ datasets and $12$ model architectures, as summarized in Appendix~\ref{appendix:models_and_datasets}. The total computation for results reported in this paper is about $3400$ hours of V100 GPU (32GB) counted by our internal computing infrastructure. More details are given in Appendix~\ref{sec:compute_estimation}.

Extensive experiments confirm the benefit of our method, and convince the community of PyTorch and MMCV/MMEngine~\citep{mmcv, mmengine2022} to quickly integrate our method. Consequently, anyone using either library can enjoy the benefit of our algorithm with as simple as a one-line code change. We provide guidelines for each library to turn on the Tune mode in Appendix~\ref{appendix:integration}.

Since Tune mode improves efficiency over Eval mode, our experiments focus on the efficiency and accuracy comparison between Eval mode and Tune mode. Train mode is often inferior in efficiency and Deploy mode is often inferior in accuracy, so we don't include them in the main paper but leave the full comparison among four modes in Appendix ~\ref{sec:original_data_tllib} and ~\ref{appendix:object_detection_details}. 

\subsection{Object Classification in Transfer Learning}
\label{sec:classification}

We first show the benefit of Tune mode in object classification in transfer learning. We use a popular open-source TLlib~\citep{jiang2022transferability}, and the datasets include CUB-200~\citep{wah2011caltech} for fine-grained bird classification, Standford Cars~\citep{krause_3d_2013} and Aircrafts~\citep{maji2013fine}. The network backbone is ResNet50 pre-trained on ImageNet. Each experiment is repeated three times with different random seeds to report mean and standard deviation. Results are reported in Table~\ref{tbl:tllib_acc}, with further details in Appendix~\ref{sec:original_data_tllib}. Compared with Eval mode, the proposed Tune mode \emph{reduces more than $7\%$ computation time and $36\%$ memory footprint}.

\begin{table}[h]
    \vspace{-10pt}
\caption{Results for Tune mode in classification using TLlib.}
\centering
\begin{tabular}{cccll}
\toprule
Dataset       & mode  & Accuracy & \tabincell{c}{Memory (GB)}            & \tabincell{c}{Time (second/iteration)} \\
\midrule
\multirow{2}{*}{CUB-200} & Eval & 82.62 ($\pm$ 0.14)    & 19.499           & 0.549             \\
              & Tune  & 83.20 ($\pm$ 0.00)    & 12.323 (\textbf{36.80\%}$\downarrow$) & 0.501 (\textbf{8.74\%}$\downarrow$)   \\
\midrule
\multirow{2}{*}{Aircrafts}     & 
Eval & 85.21 ($\pm$ 0.22)    & 19.497           & 0.548          \\
              & Tune  & 85.90 ($\pm$ 0.26)    & 12.321 (\textbf{36.81\%}$\downarrow$) & 0.505 (\textbf{7.85\%}$\downarrow$)   \\
\midrule
\multirow{2}{*}{Stanford Cars} & Eval & 90.11 ($\pm$ 0.03)    & 19.499           & 0.541            \\
              & Tune  & 90.13 ($\pm$ 0.12)    & 12.321 (\textbf{36.81\%}$\downarrow$) & 0.491 (\textbf{9.24\%}$\downarrow$)   \\
\bottomrule
\end{tabular}
\label{tbl:tllib_acc}
\end{table}

\begin{table}[ht]
\centering
\vspace{-10pt}
\caption{Object Detection results on different detectors and backbones. }
\resizebox{\textwidth}{!}{
\begin{tabular}{cccccll}
\toprule
Detector & Backbone                 & BatchSize & Precision  & mode & mAP    & Memory (GB)        \\
\midrule
\multirow{2}{*}{Faster RCNN} & \multirow{2}{*}{ResNet50}   & \multirow{2}{*}{2} & \multirow{2}{*}{FP32} & Eval & 0.3739 & 3.857              \\
                                                                                                  & & & & Tune & 0.3728 (-0.0011) & 3.003 (\textbf{22.15\%}$\downarrow$)    \\
\midrule                                        
\multirow{2}{*}{Mask RCNN}& \multirow{2}{*}{ResNet50}     & \multirow{2}{*}{2} & \multirow{2}{*}{FP32} & Eval & 0.3824 & 4.329          \\
                                                                                                  & & & & Tune & 0.3825 (+0.0001) & 3.470 (\textbf{19.85\%}$\downarrow$) \\
\midrule                                        
\multirow{2}{*}{Mask RCNN}& \multirow{2}{*}{ResNet101}    & \multirow{2}{*}{16} & \multirow{2}{*}{FP16} & Eval & 0.3755 & 13.687  \\
                                                                                                  & & & & Tune & 0.3756 (+0.0001) & 9.980 (\textbf{27.08\%}$\downarrow$) \\
\midrule
\multirow{2}{*}{Retina Net} & \multirow{2}{*}{ResNet50}    & \multirow{2}{*}{2} & \multirow{2}{*}{FP32} & Eval & 0.3675 & 3.631          \\
                                                                                                  & & & & Tune & 0.3647 (-0.0028) & 2.774 (\textbf{23.59\%}$\downarrow$) \\
\midrule                                        
\multirow{2}{*}{Faster RCNN} & \multirow{2}{*}{ResNet101}  & \multirow{2}{*}{2} & \multirow{2}{*}{FP32} & Eval & 0.3944 & 5.781          \\
                                                                                                  & & & & Tune & 0.3921 (-0.0023) & 4.183 (\textbf{27.65\%}$\downarrow$) \\
\midrule                                        
\multirow{2}{*}{Faster RCNN} & \multirow{2}{*}{ResNext101} & \multirow{2}{*}{2} & \multirow{2}{*}{FP32} & Eval & 0.4126 & 6.980          \\
                                                                                                  & & & & Tune & 0.4131 (+0.0005) & 4.773 (\textbf{31.62\%}$\downarrow$) \\
\midrule                                        
\multirow{2}{*}{Faster RCNN} & \multirow{2}{*}{RegNet}     & \multirow{2}{*}{2} & \multirow{2}{*}{FP32} & Eval & 0.3985 & 4.361          \\
                                                                                                  & & & & Tune & 0.3995 (+0.0010) & 3.138 (\textbf{28.06\%}$\downarrow$) \\
\midrule                                        
\multirow{2}{*}{Faster RCNN} & \multirow{2}{*}{HRNet}      & \multirow{2}{*}{2} & \multirow{2}{*}{FP32} & Eval & 0.4017 & 8.504          \\
                                                                                                  & & & & Tune & 0.4031 (+0.0014) & 5.463 (\textbf{35.76\%}$\downarrow$) \\
                                                                                                  \midrule                                        
\multirow{2}{*}{\href{}{Faster RCNN}} & \multirow{2}{*}{RepVGG}      & \multirow{2}{*}{16} & \multirow{2}{*}{FP16} & Eval & 0.3350 & 15.794          \\
                                                                                                  & & & & Tune & 0.3350 (+0.0000) & 8.996 (\textbf{43.04\%}$\downarrow$) \\
\bottomrule
\vspace{-15pt}
\end{tabular}
}
\label{tbl:detector_res}
\end{table}

\subsection{Object Detection in Transfer Learning}
\label{sec:detection}

This section presents object detection results on the widely used COCO~\citep{lin_microsoft_2014} dataset. The MMDetection library uses Eval mode by default, and we compare the results by switching models to Tune mode. We test against various mainstream CNN backbones and detection algorithms (including Faster RCNN~\citep{ren_faster_2015}, Mask RCNN~\citep{he_mask_2017}, and Retina Net~\citep{lin_focal_2017}). Results are displayed in Table \ref{tbl:detector_res}, with additional results available in Appendix \ref{appendix:object_detection_details}. Object detection experiments are costly, and therefore we do not repeat three times to calculate mean and standard deviation. Appendix~\ref{sec:detection_randomness} shows that the standard deviation of performance across different runs is as small as $0.0005$. The change of mAP in Table \ref{tbl:detector_res} falls into the range of random fluctuation across experiments.

With different architecture, batch size and training precision~\citep{micikevicius_mixed_2018}, \emph{Tune mode has almost the same mAP as Eval mode, while remarkably reducing the memory footprint by about $20\% \sim 40\%$}. Note that detection models typically have a pre-trained backbone for extracting features, and a head trained from scratch for producing bounding boxes and classification. The head consumes the major computation time, and the backbone consumes the major memory footprint. Because ConvBN blocks mainly lie in the backbone, our Tune mode mainly benefits the backbone, therefore reducing only the memory footprint. Computation speedup is not obvious in objection detection, and we only report the reduction of memory footprint here.


\subsection{Application of Tune Mode Beyond Transfer Learning}

Our method is designed for transfer learning. However, we find that its application can go beyond transfer learning. Any model using Eval mode can benefit from our Tune mode. Adversarial example generation~\citep{szegedy2013intriguing} is a representative application of our method: when generating adversarial examples for adversarial training~\citep{goodfellow_explaining_2015}, an important step is to calculate the gradient $\nabla_{{x}} \mathcal{L}(\theta, x, y)$ with respect to inputs $x$, given inputs, labels $y$, and parameters $\theta$. Common techniques for producing  adversarial samples, such as FGSM~\citep{goodfellow_explaining_2015}, BIM~\citep{kurakin2016adversarial}, and PGD~\citep{madry2017towards}, all perturb the inputs based on the gradients, where the model is in Eval mode. Turning on Tune mode can improve the efficiency of adversarial sample generation. Concretely, we perform forward and backward propagation of samples through the model to compute the gradient of input, and measure the time cost as well as GPU memory footprint. The experimental results can be found in Figure~\ref{fig:cost_adv}, with detailed numbers available in Appendix~\ref{appendix:cost_adv}. Across different models, Tune mode can \emph{achieve 5\%-8\% speedup and save 30\%-45\% of GPU memory.} These experiments cover widely used network architectures, and also cover UNet that has transposed convolution layers, demonstrating the broad application of our method.


\begin{figure}[h]
    \vspace{-10pt}
    \includegraphics[width=0.5\columnwidth]{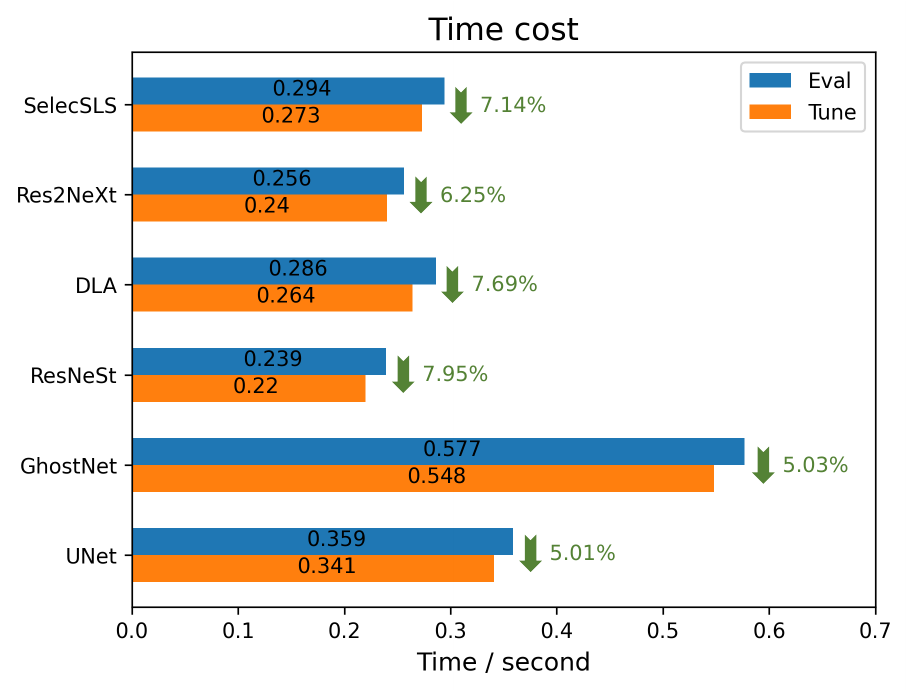}
    \includegraphics[width=0.5\columnwidth]{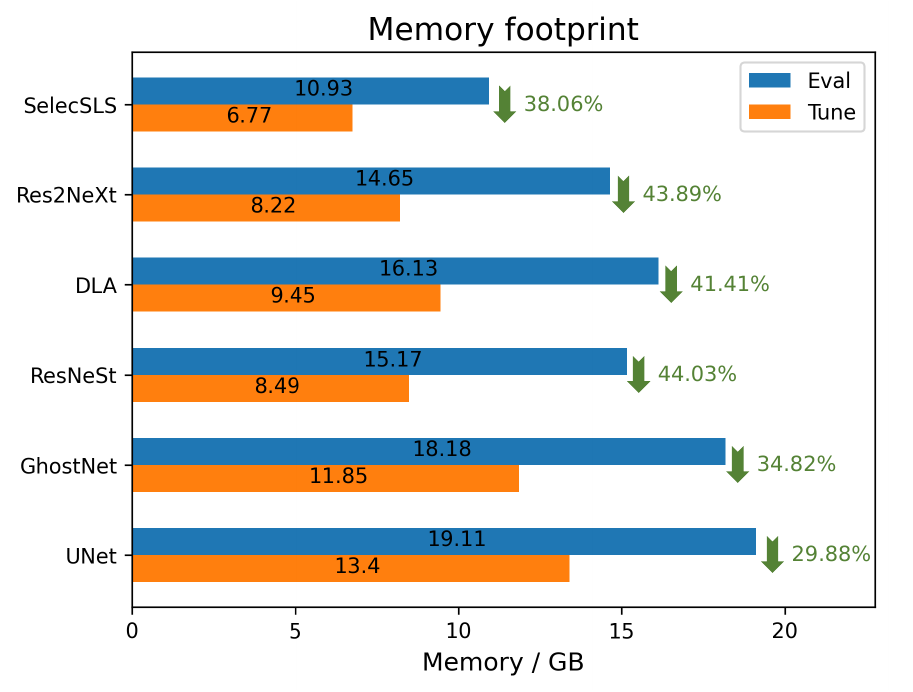}
    \vspace{-20pt}
    \caption{Tune mode \emph{v.s.} Eval mode in adversarial example generation. }
    \label{fig:cost_adv}
\end{figure}

\section{Conclusion}

This paper proposes efficient ConvBN blocks with a novel Tune mode for transfer learning and beyond. Tune Mode is equivalent to Eval mode in both forward and backward calculation while reducing memory footprint and computation time without hurting performance. Our experiments confirm the benefit across dozens of models and tasks, reducing at most $44\%$ memory footprint and $9\%$ computation time. We further bring the proposed method into open-source frameworks the community uses everyday, reducing the cost of training networks with ConvBN blocks.

\section*{Acknowledgments}
We would like to thank many open-source contributors for helping the adoption of this technique into PyTorch, MMDetection, and MMCV, including Jason Ansel from Meta, and Wenwei Zhang, Haochen Ye, Zaida Zhou from OpenMMLab.

This work was supported by the National Key Research and Development Plan (2021YFB1715200), the National Natural Science Foundation of China (U2342217 and 62022050), the BNRist Innovation Fund (BNR2024RC01010), and the National Engineering Research Center for Big Data Software.

Kaichao You is partly supported by the Apple Scholar in AI/ML.

\bibliography{iclr2024_conference}
\bibliographystyle{iclr2024_conference}

\clearpage
\appendix

\section{Comparison of Train and Eval for Object Detection}
\label{appendix:compare_train_eval}

\begin{figure}[h]
    \centering
    \vspace{-10pt}
    \includegraphics[scale=0.4]{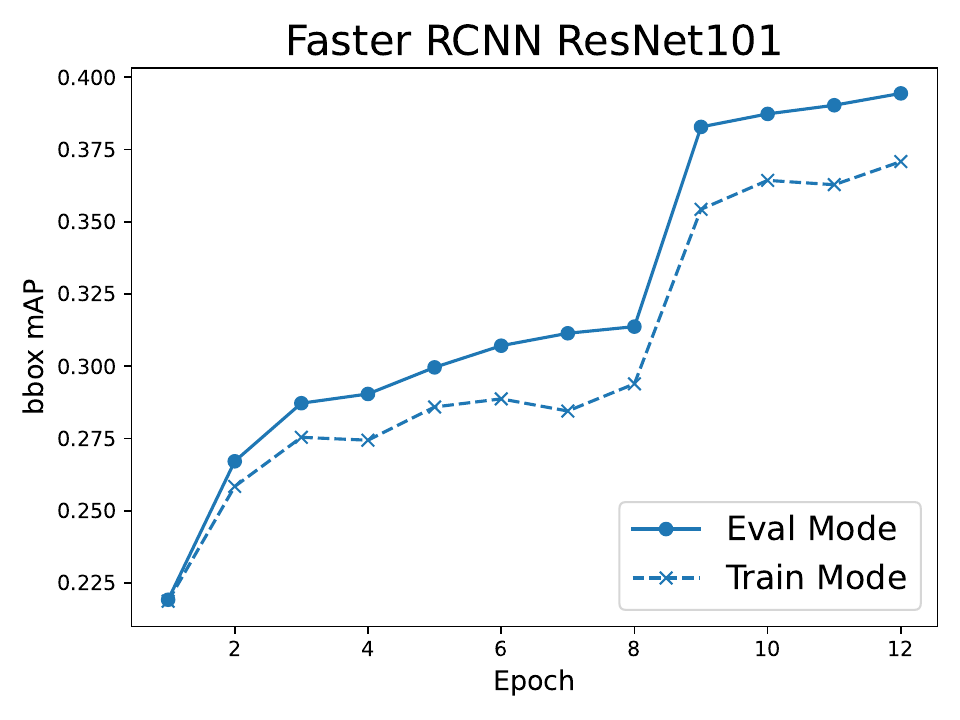}
    \hspace{0.3in}
    \includegraphics[scale=0.4]{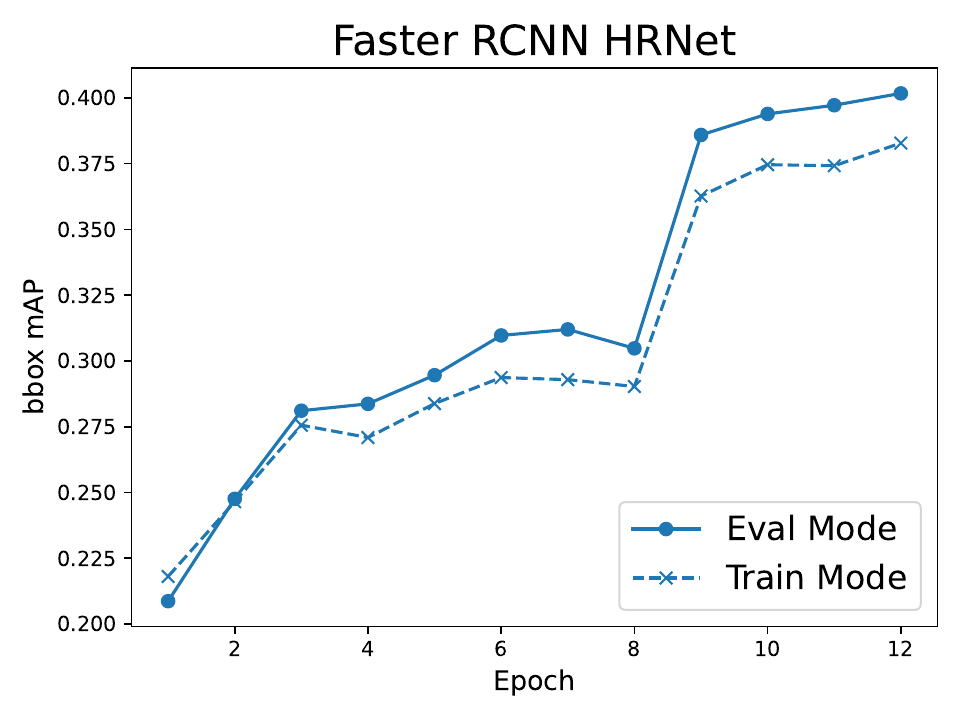}
    \vspace{-10pt}
    \caption{Training curve of Faster RCNN with \emph{ResNet101} and \emph{HRNet} backbone. Models trained in Train Mode shows noticeable performance deterioration compared to Eval Mode.}
    \vspace{-10pt}
    \label{fig:compare_eval}
\end{figure}

We compared the performance of detection models trained in Train Mode and Eval Mode, using two backbones (Resnet101 and HRNet). Results are shown in Table~\ref{tbl:compare_eval} and the training curves are shown in Figure~\ref{fig:compare_eval}.

To ensure fair comparison, we use official training schemes from MMDetection: \href{https://github.com/open-mmlab/mmdetection/blob/main/configs/faster_rcnn/faster-rcnn_r50_fpn_1x_coco.py}{faster\_rcnn/faster\_rcnn\_r50\_fpn\_1x\_coco.py} and \href{https://github.com/open-mmlab/mmdetection/blob/main/configs/hrnet/faster-rcnn_hrnetv2p-w32-1x_coco.py}{hrnet/faster\_rcnn\_hrnetv2p\_w32\_1x\_coco.py}. The default choice in MMDetection is training with Eval Mode, and we only change \verb|model.backbone.norm_eval=False| to switch training to Train Mode.

These results indicate that \emph{Eval Mode sometimes outperforms Train Mode in transfer learning of object detection}.

\begin{table}[h]
\centering
\caption{mAP of Faster RCNN trained under different ConvBN block Mode.}
\label{tbl:compare_eval}
\begin{tabular}{lll}
\toprule
Configuration File & Eval Mode & Train Mode  \\
\midrule
faster\_rcnn/faster\_rcnn\_r50\_fpn\_1x\_coco.py & 0.3944    & 0.3708      \\
hrnet/faster\_rcnn\_hrnetv2p\_w32\_1x\_coco.py & 0.4017    & 0.3828      \\
\bottomrule \\
\end{tabular}

\end{table}

\section{Backward Propagation of Broadcast}
\label{sec:broadcast}

Take the convolution operation as an example: the convolutional output $Y$ has a shape of $[N, C_\text{out}, H_\text{out}, W_\text{out}]$, while the tracked mean $\hat{\mu}$ has a shape of $[C_\text{out}]$, and $Y - \hat{\mu}$ implies first replicating $\hat{\mu}$ to have a shape of $[N, C_\text{out}, H_\text{out}, W_\text{out}]$, then performing element-wise subtraction. This can be explained by introducing an additional broadcast operator $\mathcal{B}_Y$, where $\mathcal{B}_Y(\hat{\mu})$ broadcasts $\hat{\mu}$ to match the shape of $Y$. The underlying calculation is actually $Y - \mathcal{B}_Y(\hat{\mu})$. The backward calculation for the broadcast operator $\pd{\mathcal{B}_Y(\hat{\mu})}{\hat{\mu}}$ is the reverse of replication, i.e., summing over a large tensor with the shape of $[N, C_\text{out}, H_\text{out}, W_\text{out}]$ into a smaller tensor with the shape of $[C_\text{out}]$. This helps understand the backward equation for the bias term $\pd{J}{b} = \pd{J}{Y}$, which actually means $\pd{J}{b} = \pd{\mathcal{B}_Y(b)}{b} \pd{J}{Y}$, \emph{i.e.}, summing $\pd{J}{Y}$ to be compatible with the shape of $\pd{J}{b}$. Due to the prevalence of broadcast in neural networks, we omit them to simplify equations. 

\clearpage
\newpage

\section{Code Details of Train/Eval/Deploy Mode}
\label{appendix:code_detail_3_mode}

Cmputation details of ConvBN blocks in different modes, with shape annotations for each tensor available in the following code snippet.

\begin{lstlisting}[float=h, caption=Computation details for consecutive Convolution and BatchNorm layers in different modes, label=lst:sample_code]
# input: A faeture map (*@$X$@*). The shape of (*@$X$@*) is (*@$[N, C_{in}, H_{in}, W_{in}]$@*).
# input: A convolutional layer "conv" with kernel-size (*@$k$@*) and output channel number (*@$C_{out}$@*); it has a weight parameter (*@$W$@*) with the shape of (*@$[C_{out}, C_{in}, k, k]$@*) and a bias parameter (*@$b$@*) with the shape of (*@$[C_{out}]$@*).
# input: A BatchNorm layer "bn" with output channel number (*@$C_{out}$@*); it has a weight parameter (*@$\gamma$@*) with the shape of (*@$[C_{out}]$@*), and a weight parameter (*@$\beta$@*) with the shape of (*@$[C_{out}]$@*). The momentum update rate of BatchNorm is a constant number (*@$\alpha \in (0,1)$@*).
# output: Z = bn(conv(X))

# code explained in a pytorch style
import torch

"Train Mode"
# calculate the output of convolution
(*@$Y$@*) = (*@$W \circledast\ X + b$@*) # (*@$\circledast$@*) for convolution. The shape of (*@$Y$@*) is (*@$[N, C_{out}, H_{out}, W_{out}]$@*)
# calculate the mean for normalization
(*@$\mu$@*) = torch.mean((*@$Y$@*), dim=(0, 2, 3)) # (*@$\mu$@*) has a shape of (*@$[C_{out}]$@*)
# calculate the variation for normalization
(*@$\sigma^2$@*) = torch.var((*@$Y$@*), dim=(0, 2, 3)) # (*@$\sigma^2$@*) has a shape of (*@$[C_{out}]$@*)
# update tracked statistics, (*@$\hat{\mu}$@*) and (*@$\hat{\sigma^2}$@*) keep track of moving mean and moving variance. They are initialized to 0 and 1 respectively if the model is trained from scratch, or are inherited from pre-trained values.
(*@$\hat{\mu} \leftarrow \hat{\mu} + \alpha (\mu-\hat{\mu})$@*)
(*@$\hat{\sigma}^2 \leftarrow \hat{\sigma}^2 + \alpha (\sigma^2-\hat{\sigma}^2)$@*)

# normalize the output
# (*@$\epsilon$@*) is a small positive number to avoid zero division
(*@$\bar{Y} = \frac{Y - \mu}{\sqrt{\sigma^2 + \epsilon}}$@*) # (*@$\mu$@*) and (*@$\sigma^2$@*) are broadcast to match the shape of (*@$Y$@*)
# apply the affine transform
(*@$Z = \gamma * \bar{Y} + \beta$@*) # (*@$\gamma$@*) and (*@$\beta$@*) are broadcast to match the shape of (*@$\bar{Y}$@*)

"Eval Mode"
# calculate the output of convolution
(*@$Y$@*) = (*@$W \circledast X + b$@*)
# normalize the output with tracked statistics
(*@$\bar{Y} = \frac{Y - \hat{\mu}}{\sqrt{\hat{\sigma^2} + \epsilon}}$@*)
# apply the affine transform
(*@$Z = \gamma * \bar{Y} + \beta$@*)

"Deploy Mode"
# update the weight and bias of the convolution once for all
(*@$\hat{W} = W * \frac{\gamma}{\sqrt{\hat{\sigma}^2 + \epsilon}}$@*)
(*@$\hat{b} = (b - \hat{\mu})\frac{\gamma}{\sqrt{\hat{\sigma}^2 + \epsilon}} + \beta$@*)

# convolution with updated parameters is equivalent to consecutive convolution and batch normalization 
(*@$Z$@*) = (*@$\hat{W} \circledast X + \hat{b}$@*)

\end{lstlisting}
\clearpage
\newpage

\section{Efficiency Comparison Between Tune and Deploy}
\label{sec:tune_and_deploy_compare}

\begin{table}[h]
    \centering
    \caption{Efficiency comparison between Eval, Tune and Deploy. Time is measured by second/iteration and the memory is the peak memory footprint (GB) during forward and backward. We can observe that $\text{Deploy} \approx \text{Tune} > \text{Eval}$ in terms of efficiency.}
    \begin{tabular}{cccccccccc}
    \toprule
    \multirow{2}{*}{\tabincell{c}{Batch \\ Size}} & \multirow{2}{*}{\tabincell{c}{Input \\ Size}} & \multicolumn{2}{c}{Eval Mode} & \multicolumn{2}{c}{Tune Mode} & \multicolumn{2}{c}{DeployMode} \\
                                &                             & Time          & Memory        & Time          & Memory        & Time          & Memory  \\
    \midrule
    32  & 224 & 0.0945 & 2.8237  & 0.0849 & 1.5973  & 0.0830 & 1.5619    \\
    32  & 256 & 0.1110 & 3.5965  & 0.1032 & 1.9732  & 0.1011 & 1.9416  \\
    32  & 288 & 0.1488 & 4.5130  & 0.1382 & 2.4325  & 0.1356 & 2.3728   \\
    32  & 320 & 0.1761 & 5.5216  & 0.1630 & 2.9207  & 0.1609 & 2.8363   \\
    32  & 352 & 0.2153 & 6.6120  & 0.1991 & 3.4546  & 0.1969 & 3.3682   \\
    32  & 384 & 0.2503 & 7.9005  & 0.2304 & 4.0995  & 0.2281 & 4.0219  \\
    32  & 416 & 0.2983 & 9.2317  & 0.2738 & 4.7429  & 0.2721 & 4.6640  \\
    32  & 448 & 0.3567 & 10.6448 & 0.3104 & 5.4306  & 0.3077 & 5.3421   \\
    16  & 224 & 0.0505 & 1.5671  & 0.0467 & 0.9727  & 0.0448 & 0.9397  \\
    32  & 224 & 0.0948 & 2.8237  & 0.0849 & 1.5973  & 0.0831 & 1.5631   \\
    64  & 224 & 0.1837 & 5.4125  & 0.1613 & 2.8617  & 0.1590 & 2.7808  \\
    128 & 224 & 0.3577 & 10.6001 & 0.3081 & 5.4088  & 0.3060 & 5.3284  \\
    256 & 224 & 0.7035 & 21.0107 & 0.6001 & 10.5011 & 0.5966 & 10.4234 \\
    \bottomrule
    \end{tabular}
\end{table}

\section{Models and Datasets Tested}
\label{appendix:models_and_datasets}
We conduct extensive experiments in object detection, classification, and adversarial example generation.

Our experiments cover $5$ datasets:

\begin{itemize}
    \item CUB-200~\citep{wah2011caltech}: CUB-200 is the most widely-used dataset for fine-grained visual categorization task. It contains 11,788 images of 200 subcategories belonging to birds.
    \item Standford Cars~\citep{krause_3d_2013}: Standford Cars consists of 196 classes of cars with a total of 16,185 images. 
    \item Aircrafts~\citep{maji2013fine}: Aircrafts contains 10,200 images of aircraft, with 100 images for each of 102 different aircraft model variants. 
    \item COCO~\citep{lin_microsoft_2014}: COCO is a large-scale object detection, segmentation, key-point detection, and captioning dataset released by Microsoft. The dataset consists of 328K images.
    \item ImageNet~\citep{deng_imagenet:_2009}: ImageNet dataset contains 14,197,122 annotated images according to the WordNet hierarchy and it is instrumental in advancing computer vision and deep learning research.
\end{itemize}

Our experiments cover $12$ model architectures:

\begin{itemize}
    \item ResNet50, ResNet101~\citep{he_deep_2016}: ResNet introduced the residual structure, making it possible to train models with hundreds or thousands of layers, which was a significant breakthrough in deep learning.
    \item ResNeXt101~\citep{xie2017aggregated}: ResNeXt is constructed by repeating a building block that aggregates a set of transformations with the same topology and achieved second place in ILSVRC 2016.
    \item RegNet~\citep{radosavovic2020designing}: RegNet is a self-regulated network for image classification and can be easily implemented and appended to any ResNet architecture.
    \item HRNet~\citep{sun2019deep}: HRNet is a general purpose convolutional neural network for tasks like semantic segmentation, object detection and image classification and is able to maintain high resolution representations through the whole process. 
    \item RepVGG~\citep{ding_repvgg:_2021}: RepVGG is a simple but powerful architecture of convolutional neural network, which has a VGG-like inference-time body composed of nothing but a stack of 3x3 convolution and ReLU.
    \item SelecSLS~\citep{mehta2020xnect}: SelecSLS uses novel selective long and short range skip connections to improve the information flow allowing for a drastically faster network without compromising accuracy.
    \item Res2NeXt~\citep{gao_res2net:_2019}: Res2NeXt represents multi-scale features at a granular level and increases the range of receptive fields for each network layer.
    \item DLA~\citep{yu2018deep}: Extending “shallow” skip connections, DLA incorporates more depth and sharing. It contains iterative deep aggregation (IDA) and hierarchical deep aggregation (HDA).
    \item ResNeSt~\citep{zhang2022resnest}: ResNeSt applies the channel-wise attention on different network branches to leverage their success in capturing cross-feature interactions and learning diverse representations.
    \item GhostNet~\citep{han_ghostnet:_2020}: GhostNet is a type of convolutional neural network that is built using Ghost modules, which aim to generate more features by using fewer parameters.
    \item UNet~\citep{ronneberger2015u}: UNet consists of a contracting path and an expansive path and is widely employed across various facets of semantic segmentation.
\end{itemize}

\section{Estimation of Total Computation Used in This Paper}
\label{sec:compute_estimation}

Each trial of classification experiments in Section~\ref{sec:classification} requires about $2$ hours of V100 GPU training. The numbers reported in this paper requires 18 trials, which cost about $36$ GPU hours.

Each trial of detection experiments in Section~\ref{sec:detection} requires about $12$ hours of 8 V100 GPU training, which is $96$ GPU hours. The numbers reported in this paper requires 25 trials, which cost about $2400$ GPU hours.

Each trial of pre-training experiments in Section~\ref{sec:pre-training} requires about $24$ hours of 8 V100 GPU training, which is $192$ GPU hours. The numbers reported in this paper requires about $\frac{5}{3}$ full trials, which cost about $320$ GPU hours.

Some experiments for the purpose of analyses also cost computation. Figure~\ref{fig:coeff_distribution_and_deploy_mode_failure} requires two trials of object detection and 6 trials of object classification, with about $204$ GPU hours. Figure~\ref{fig:compare_eval} requires four trials of object detection, with about $384$ GPU hours.

Summing the above numbers up, and considering all the fractional computation for the rest analyses experiments, \emph{this paper costs about $3400$ GPU hours}.

Considering the cost of prototyping and previous experiments that do not get into the paper, \emph{the total cost of this project is about $5000$ GPU hours}.

Note that these numbers are rough estimation of the cost, and do not include the additional cost for storing data/system maintenance etc.

\section{Comparison of Four Modes for Object Classification}
\label{sec:original_data_tllib}

The below settings are taken from the default values in the TLlib library: ResNet50 is the backbone network and all parameters are optimized by Stochastic Gradient Descent with 0.9 momentum and 0.0005 weight decay. Each training process consisted of 20 epochs, with 500 iterations per epoch. We set the initial learning rates to 0.001 and 0.01 for the feature extractor and linear projection head respectively, and scheduled the learning rates of all layers to decay by 0.1 at epochs 8 and 12. The input images were all resized and cropped to 448 $\times$ 448, and the batch size was fixed at 48. Since the backbone network takes the major computation, the memory and time in three different dataset are very similar.
\clearpage
\newpage

\begin{table}[h]
\caption{Comparison of four modes in classification using TLlib.}
\centering
\begin{tabular}{cccll}
\toprule
Dataset       & mode  & Accuracy & \tabincell{c}{Memory (GB)}            & \tabincell{c}{Time (second/iteration)} \\
\midrule
\multirow{4}{*}{CUB-200} & Train & 83.07     & 19.967           & 0.571             \\
& Eval & 82.62 & 19.499 & 0.549 
\\
& Deploy &  62.96 & 12.002 &  0.511
\\
              & Tune  & 83.20     & 12.323  & 0.501   \\
              
\midrule
\multirow{4}{*}{Aircrafts}     & Train & 85.40  & 19.965           & 0.564             \\
& Eval & 85.21 & 19.497 &  0.548 \\
& Deploy & 41.22 & 12.000  & 0.506 
\\
              & Tune  & 85.90    & 12.321 & 0.505  \\
\midrule
\multirow{4}{*}{Stanford Cars} & Train & 89.87    & 19.967           & 0.571             \\
& Eval &  90.11 &  19.499  &  0.541 
\\
& Deploy & 47.42 & 12.002 & 0.507 
\\
              & Tune  & 90.13  & 12.321 & 0.491  \\
\bottomrule
\end{tabular}
\label{tbl:appendix_tllib_acc}
\end{table}

\section{Detailed Object Detection Experimental Results}
\label{appendix:object_detection_details}
For object detection, more detailed comparison of Eval mode and Tune mode is presented in Table ~\ref{tbl:appendix_detector_res}, while a comparison of the four modes can be found in Table ~\ref{tbl:appendix_four_mode_det}.

\begin{table}[htbp]
\centering
\caption{Detailed Object Detection experimental results.}
\resizebox{\textwidth}{!}{
\begin{tabular}{cccccll}
\toprule
Detector & Backbone                 & BatchSize & Precision  & mode & mAP    & Memory(GB)        \\
\midrule
\multirow{2}{*}{Faster RCNN} & \multirow{2}{*}{ResNet50}   & \multirow{2}{*}{2} & \multirow{2}{*}{FP32} & Eval & 0.3739 & 3.857              \\
                                                                                                  & & & & Tune & 0.3728 (-0.0011) & 3.003 (\textbf{22.15\%}$\downarrow$)    \\
\midrule                                        
\multirow{2}{*}{Mask RCNN}& \multirow{2}{*}{ResNet50}     & \multirow{2}{*}{2} & \multirow{2}{*}{FP32} & Eval & 0.3824 & 4.329          \\
                                                                                                  & & & & Tune & 0.3825 (+0.0001) & 3.470 (\textbf{19.85\%}$\downarrow$) \\
\midrule                                        
\multirow{2}{*}{Mask RCNN}& \multirow{2}{*}{ResNet101}    & \multirow{2}{*}{16} & \multirow{2}{*}{FP16} & Eval & 0.3755 & 13.687  \\
                                                                                                  & & & & Tune & 0.3756 (+0.0001) & 9.980 (\textbf{27.08\%}$\downarrow$) \\
\midrule
\multirow{2}{*}{Retina Net} & \multirow{2}{*}{ResNet50}    & \multirow{2}{*}{2} & \multirow{2}{*}{FP32} & Eval & 0.3675 & 3.631          \\
                                                                                                  & & & & Tune & 0.3647 (-0.0028) & 2.774 (\textbf{23.59\%}$\downarrow$) \\
\midrule                                        
\multirow{2}{*}{Faster RCNN} & \multirow{2}{*}{ResNet101}  & \multirow{2}{*}{2} & \multirow{2}{*}{FP32} & Eval & 0.3944 & 5.781          \\
                                                                                                  & & & & Tune & 0.3921 (-0.0023) & 4.183 (\textbf{27.65\%}$\downarrow$) \\
\midrule                                        
\multirow{2}{*}{Faster RCNN} & \multirow{2}{*}{ResNet101}  & \multirow{2}{*}{2} & \multirow{2}{*}{FP16} & Eval & 0.3944 & 3.849          \\
                                                                                                  & & & & Tune & 0.3925 (-0.0019) & 3.138 (\textbf{18.47\%}$\downarrow$) \\
\midrule
\multirow{2}{*}{Faster RCNN} & \multirow{2}{*}{ResNet101}  & \multirow{2}{*}{8} & \multirow{2}{*}{FP16} & Eval & 0.3922 & 10.411   \\
                                                                                                  & & & & Tune & 0.3917 (-0.0005) & 7.036 (\textbf{32.41\%}$\downarrow$) \\
\midrule
\multirow{2}{*}{Faster RCNN} & \multirow{2}{*}{ResNet101}  & \multirow{2}{*}{16} & \multirow{2}{*}{FP16} & Eval & 0.3902 & 19.799 \\
                                                                                                  & & & & Tune & 0.3899 (-0.0003) & 12.901(\textbf{34.83\%}$\downarrow$) \\
\midrule
\multirow{2}{*}{Faster RCNN} & \multirow{2}{*}{ResNext101} & \multirow{2}{*}{2} & \multirow{2}{*}{FP32} & Eval & 0.4126 & 6.980          \\
                                                                                                  & & & & Tune & 0.4131 (+0.0005) & 4.773 (\textbf{31.62\%}$\downarrow$) \\
\midrule                                        
\multirow{2}{*}{Faster RCNN} & \multirow{2}{*}{RegNet}     & \multirow{2}{*}{2} & \multirow{2}{*}{FP32} & Eval & 0.3985 & 4.361          \\
                                                                                                  & & & & Tune & 0.3995 (+0.0010) & 3.138 (\textbf{28.06\%}$\downarrow$) \\
\midrule                                        
\multirow{2}{*}{Faster RCNN} & \multirow{2}{*}{HRNet}      & \multirow{2}{*}{2} & \multirow{2}{*}{FP32} & Eval & 0.4017 & 8.504          \\
                                                                                                  & & & & Tune & 0.4031 (+0.0014) & 5.463 (\textbf{35.76\%}$\downarrow$) \\
                                                                                                  \midrule                                        
\multirow{2}{*}{\href{}{Faster RCNN}} & \multirow{2}{*}{RepVGG}      & \multirow{2}{*}{16} & \multirow{2}{*}{FP16} & Eval & 0.3350 & 15.80          \\
                                                                                                  & & & & Tune & 0.3350 (+0.0000) & 9.00 (\textbf{43.04\%}$\downarrow$) \\
\bottomrule
\vspace{-15pt}
\end{tabular}
}
\label{tbl:appendix_detector_res}
\end{table}
\clearpage
\newpage

\begin{table}[h]
\caption{Comparison of four modes in detection.}
\centering
\resizebox{\textwidth}{!}{
\begin{tabular}{ccccllll}
\toprule
Detector       & Backbone  & Batchsize & Precision     & mode & mAP  & Memory(GB) & Time(sec/iter) \\
\midrule
\multirow{4}{*}{Faster RCNN} & \multirow{4}{*}{ResNet101} & \multirow{4}{*}{2} & \multirow{4}{*}{FP32} &Train & 0.3708    & 5.782  &  0.3116           \\
& & & &  Eval & 0.3944&   5.781  & 0.3060
\\
& & & & Deploy & 0.3690 & 4.02   &  0.3060 
\\
& & & & Tune  &0.3921  & 4.18 & 0.3085 \\
\bottomrule
\end{tabular}
}
\label{tbl:appendix_four_mode_det}
\end{table}

\section{Small Randomness in Object Detection}
\label{sec:detection_randomness}

The high computational cost limited us to repeating the experiment only once for validating small randomness in object detection. We conducted three trials of the \href{https://github.com/open-mmlab/mmdetection/blob/main/configs/faster_rcnn/faster-rcnn_r50_fpn_1x_coco.py}{Faster RCNN ResNet50} standard configuration and obtained an average best mAP of 0.3748, 0.3735, and 0.3739, with a standard deviation of 0.000543. These results demonstrate that object detection tasks have very little randomness.

\section{Adversarial Example Generation}
\label{appendix:cost_adv}
Table ~\ref{tbl:appendix_adv} presents more detailed model information and batch size information for the adversarial example generation experiments.

\begin{table}[h]
\caption{Adversarial example generation result.}
\centering
\resizebox{\textwidth}{!}{
\begin{tabular}{ccllllll}
\toprule
Arch &  & ResNeSt101e  & DLA102 & SelecSLS42b   & Res2NeXt50 & GhostNet\_100 & UNet \\
\midrule
Batch size &  & 64 & 128 & 256 & 128 & 512 & 128\\
\midrule
Time & Eval & 0.239   & 0.286 & 0.294 & 0.256 & 0.577    &  0.359    \\
(s)  & Tune  & 0.22 (\textbf{7.95}\%$\downarrow$)      & 0.264(\textbf{7.69}\%$\downarrow$)& 0.273 (\textbf{7.14}\%$\downarrow$) & 0.240 (\textbf{6.25}\%$\downarrow$) & 0.548 (\textbf{5.03}\%$\downarrow$) & 0.341(\textbf{5.01}\%$\downarrow$)\\
\midrule
Memory   & Eval & 15.17    & 16.13           & 10.93 & 14.65 & 18.18  &   19.11     \\
(GB)  & Tune  & 8.49 (\textbf{44.03}\%$\downarrow$)  & 9.45 (\textbf{41.41}\%$\downarrow$) & 6.77 (\textbf{38.06}\%$\downarrow$)  & 8.22 (\textbf{43.89}\%$\downarrow$)  & 11.85 (\textbf{34.82}\%$\downarrow$) & 13.40 (\textbf{29.88}\%$\downarrow$)   \\
\bottomrule
\end{tabular}
}
\label{tbl:appendix_adv}
\end{table}

\section{Tune Mode ConvBN for Pre-training}
\label{sec:pre-training}

Tune mode is designed for transfer learning because it requires tracked statistics to normalize features. Here we show that Tune mode can also be used in late stages of pre-training. We use the prevalent ImageNet pre-training as baseline, which has three stages with decaying learning rate. We tried to turn on Tune mode at the third stage, the accuracy slightly dropped. Nevertheless, due to our implementation with \texttt{torch.fx}, we can dynamically switch the mode during training. Therefore, we also tried to alternate between Train mode and Tune mode at the third stage, which retained the accuracy with less computation time.

\begin{figure}[h]
    \centering
    \vspace{-10pt}
    \includegraphics[width=\textwidth]{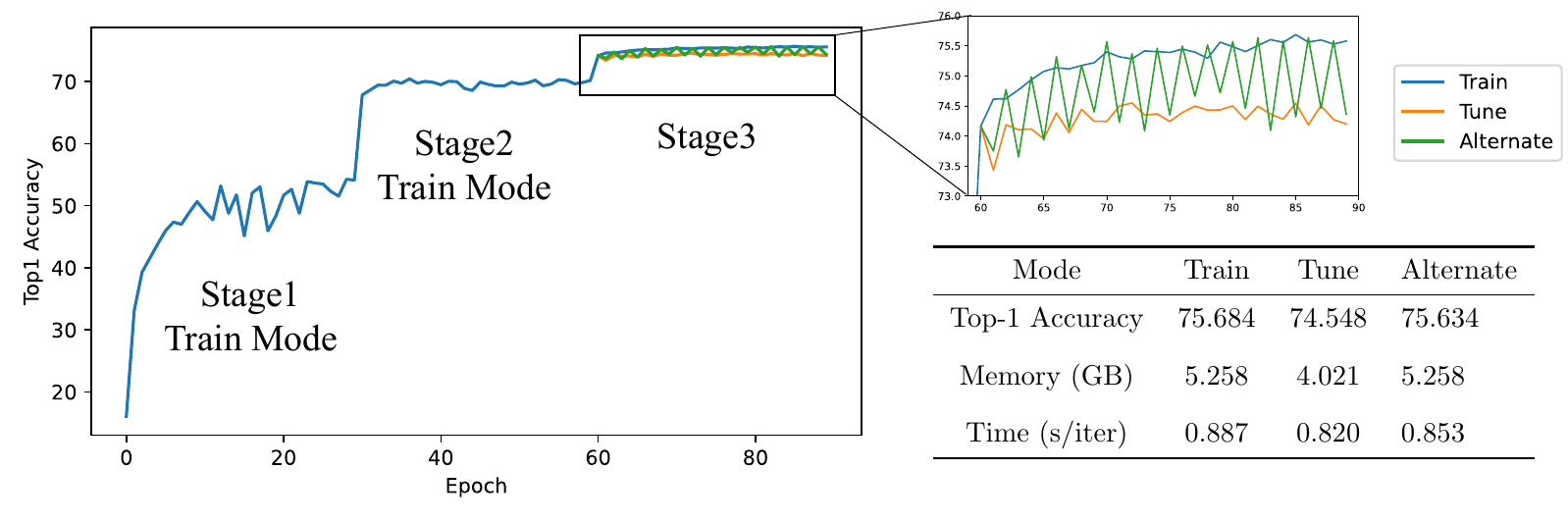}
    \vspace{-10pt}
    \caption{ImageNet Pre-training Results}
    \label{fig:imagenet}
\end{figure}

\section{Comparing with alternatives to reducing memory footprint}
\label{appendix:alternative}

\subsection{FrozenBatchNorm}

FrozenBatchNorm, as used in Detectron2~\citep{wu2019detectron2}, freezes the weight and bias of BatchNorm layers, reducing memory footprint at the cost of less trainable parameters, which limits the network's expressive power and hinders the accuracy of trained models. As shown in Table~\ref{tab:compare_frozenbn}, FrozenBatchNorm's mAP in detection is lower than baseline, while our Tune mode doesn't hurt mAP.

\begin{table}[h]
    \centering
    \caption{Comparision between proposed detection baseline and FrozenBatchNorm. FrozenBatchNorm incurs performance loss.}
    \label{tab:compare_frozenbn}
    \begin{tabular}{ccc}
    \toprule
                    & Faster RCNN ResNet 50 2X & Mask RCNN ResNet 50 2X \\
    \midrule
    Baseline        & 0.3832                   & 0.3911                 \\
    FrozenBatchNorm & 0.3790                   & 0.3904                 \\
    \bottomrule
    \end{tabular}
\end{table}

\subsection{Inplace-ABN}
We compare our proposed Tune Mode with the Inplace-ABN~\citep{bulo_-place_2018}, a memory reduction method for ConvBN blocks by invertible activations. We show transfer learning experiments on TLlib~\citep{jiang2022transferability} with the same settings of section \ref{sec:classification}.

To apply Inplace-ABN blocks, we find BN-ReLU patterns in the pretrained network using \texttt{torch.fx} and replace them with the Inplace-ABN blocks provided by \citet{bulo_-place_2018}. Results are summarized in Table \ref{tbl:inplace_abn_appendix}. Inplace-ABN block saves approximately 16\% memory at the cost of $8\%$ additional computation time, and also hurts the accuracy significantly as it modifies the network architecture. Compared to Inplace-ABN, our proposed Tune mode saves more memory footprint and requires less computation time, while retaining the accuracy.

\begin{table}[h]
    \caption{Comparision between proposed Tune Mode and the Inplace-ABN \citep{bulo_-place_2018}. We can observe that Inplace-ABN saves memory at the cost of speed and accuracy, while Tune mode saves memory at no cost of computation time or accuracy.}
    \centering
    \resizebox{\textwidth}{!}{
    \begin{tabular}{cllll}
    \toprule
                    Dataset     & Method  & Accuracy       & Memory (GB)           & Time (second/iteration)             \\
    \midrule
    \multirow{4}{*}{CUB-200}    & Baseline         & $\text{83.07}_{\pm0.15}$    & 19.967           & 0.571            \\[2mm]
                                & Inplace-ABN      & $\text{74.80}_{\pm0.21}$    (8.27$\downarrow$) & 16.739 (16.17\%$\downarrow$)           & 0.623 (9.10\%$\uparrow$)           \\[2mm]
                                & Tune Mode (ours) & $\textbf{83.20}_{\pm0.00}$  (0.13$\uparrow$)   & 12.323 (\textbf{38.28\%}$\downarrow$) & 0.501 (\textbf{12.26\%}$\downarrow$) \\
    \midrule
    \multirow{4}{*}{Aircraft}   & Baseline         & $\text{85.40}_{\pm0.20}$  & 19.965            & 0.564            \\[2mm]
                                & Inplace-ABN      & $\text{78.23}_{\pm0.45}$    (7.17$\downarrow$)& 16.737 (16.17\%$\downarrow$)           & 0.620 (8.58\%$\uparrow$)           \\[2mm]
                                & Tune Mode (ours) & $\textbf{85.90}_{\pm0.26}$  (0.50$\uparrow$)& 12.323 (\textbf{38.28}\%$\downarrow$) & 0.505 (\textbf{10.51}\%$\downarrow$) \\
    \midrule
    \multirow{4}{*}{Stanford Car}& Baseline        & $\text{89.87}_{\pm0.06}$    & 19.967           & 0.571            \\[2mm]
                                & Inplace-ABN      & $\text{86.30}_{\pm0.28}$    (3.57$\downarrow$)& 16.739 (16.17\%$\downarrow$)           & 0.614 (7.53\%$\uparrow$)           \\[2mm]
                                & Tune Mode (ours) & $\textbf{90.13}_{\pm0.12}$  (0.26$\uparrow$) & 12.321 (\textbf{38.28\%}$\downarrow$) & 0.491 (\textbf{14.00\%}$\downarrow$) \\

    \bottomrule
    \label{tbl:inplace_abn_appendix}
    \end{tabular}
    }
\end{table}

\section{Theoretical analyses of benefit in memory/time cost}
\label{appendix:theoretical}

\subsection{Memory analysis}

Memory cost for Eval mode: $\mathcal{O}(X + Y) = \mathcal{O}(N C_{\text{in}} H_{\text{in}} W_{\text{in}} + N C_{\text{out}} H_{\text{out}} W_{\text{out}})$.

Memory cost for Tune mode: $\mathcal{O}(X + \omega^\prime) = \mathcal{O}(N C_{\text{in}} H_{\text{in}} W_{\text{in}} + k^2C_{\text{in}} C_{\text{out}})$.

For each ConvBN block, \emph{the memory cost reduction of Tune mode} is $\mathcal{O}(N C_{\text{out}} H_{\text{out}} W_{\text{out}} - k^2C_{\text{in}} C_{\text{out}})$.

From the analysis, we can conclude that networks with larger feature maps (larger $H_{\text{out}}W_{\text{out}}$, such as HRNet that features high resolutions), smaller kernel sizes (smaller $k$, such as RepVGG with many $k=1$ conv kernels), and larger batch sizes (larger $N$) will benefit more from the proposed Tune mode. The conclusion can be empirically validated from Table~\ref{tbl:appendix_detector_res}. We can observe that:

\begin{itemize}
    \item The memory cost reduction ratio grows from ${18.47\%}$ to ${34.83\%}$ when batch size grows from $2$ to $16$, for the same Faster RCNN detector with ResNet101 backbone.
    \item The memory cost reduction ratio grows from ${18.47\%}$ to ${35.76\%}$ when changing the network backbone from ResNet101 to HRNet while keeping the rest the same.
    \item The memory cost reduction ratio grows from ${34.83\%}$ to ${43.04\%}$ when changing the network backbone from ResNet101 to RepVGG while keeping the rest the same.
\end{itemize}

\subsection{Time analysis}

As pointed out by the \href{https://proceedings.neurips.cc/paper_files/paper/2022/hash/67d57c32e20fd0a7a302cb81d36e40d5-Abstract-Conference.html}{FlashAttention paper}~\citep{dao_flashattention:_2022}, \emph{the computation time of modern GPU hardware often scales with memory access} (the number of bytes the program reads from and writes to memory). The cost of memory access can be effectively estimated by summing all node sizes from computation graphs in Table~\ref{tab:conv_bn_details}. Therefore, we can analyze the time cost of Eval mode and Tune mode as follows:

Eval mode time cost:

$\mathcal{O}(X + Y + \omega + b + \bar{Y} + \hat{\mu} + \hat{\sigma} + \beta + \gamma + Z) = \mathcal{O}(N C_{\text{in}} H_{\text{in}} W_{\text{in}} + 3 N C_{\text{out}} H_{\text{out}} W_{\text{out}} + k^2C_{\text{in}} C_{\text{out}} + 5 C_{\text{out}})$.

Tune mode time cost:

$\mathcal{O}(X + \omega + b + \omega^\prime + b^\prime + \hat{\mu} + \hat{\sigma} + \beta + \gamma + Z) = \mathcal{O}(N C_{\text{in}} H_{\text{in}} W_{\text{in}} + N C_{\text{out}} H_{\text{out}} W_{\text{out}} + 2 k^2C_{\text{in}} C_{\text{out}} + 6 C_{\text{out}})$.

For each ConvBN block, \emph{the time cost reduction of Tune mode} is:

$\mathcal{O}(2 N C_{\text{out}} H_{\text{out}} W_{\text{out}} - k^2C_{\text{in}} C_{\text{out}} - C_{\text{out}})$.

Since feature maps (size $N C_{\text{out}} H_{\text{out}} W_{\text{out}}$) are typically much larger than convolutional kernels (size $k^2C_{\text{in}} C_{\text{out}}$), the proposed Tune mode can reduce time cost.

\section{Integration with DL compilers and common libraries}
\label{appendix:integration}

Our algorithm has been integrated into PyTorch core, MMCV, and MMEngine. We also support standalone usage.

\subsection{PyTorch}

Our method has been integrated into PyTorch core since version 2.2. People using PyTorch can turn on the Tune mode via the following \href{https://openreview.net/forum?id=lHZm9vNm5H&noteId=cgxGu9HEox}{code}:

\begin{lstlisting}
    torch._inductor.config.efficient_conv_bn_eval_fx_passes = True
\end{lstlisting}

\subsection{MMCV/MMEngine}

Our method has been integrated into popular computer vision libraries. MMCV holds core operators, while MMEngine is the training framework. People using MMCV/MMEngine can turn on the Tune mode via adding a command line argument:

\begin{lstlisting}
    --cfg-options efficient_conv_bn_eval="[backbone]"
\end{lstlisting}

\subsection{Standalone Usage}

For people using old versions of PyTorch (we require PyTorch larger than 1.8), they can turn on Tune mode via the online \href{https://github.com/apple/ml-tune-mode-convbn}{code}.

\begin{lstlisting}
model = MyModel() # init a model
import tune_mode_convbn
tune_mode_convbn.turn_on_efficient_conv_bn_eval_for_single_model(model)
# now this model can benefit from tune mode, if it is trained with `Eval` mode.
\end{lstlisting}

\end{document}

%% file: math_commands.tex
\usepackage{amsmath,amsfonts,bm}

\def\eqref#1{equation~\ref{#1}}

\def\1{\bm{1}}

\DeclareMathAlphabet{\mathsfit}{\encodingdefault}{\sfdefault}{m}{sl}
\SetMathAlphabet{\mathsfit}{bold}{\encodingdefault}{\sfdefault}{bx}{n}













%% file: iclr2024_conference.bbl
\begin{thebibliography}{58}
\providecommand{\natexlab}[1]{#1}
\providecommand{\url}[1]{\texttt{#1}}
\expandafter\ifx\csname urlstyle\endcsname\relax
  \providecommand{\doi}[1]{doi: #1}\else
  \providecommand{\doi}{doi: \begingroup \urlstyle{rm}\Url}\fi

\bibitem[Ba et~al.(2016)Ba, Kiros, and Hinton]{ba_layer_2016}
Jimmy~Lei Ba, Jamie~Ryan Kiros, and Geoffrey~E. Hinton.
\newblock Layer {Normalization}.
\newblock In \emph{{NeurIPS} {Deep} {Learning} {Symposium} Workshop}, 2016.

\bibitem[Bishop(2006)]{bishop_pattern_2006}
Christopher~M. Bishop.
\newblock \emph{Pattern recognition and machine learning}.
\newblock 2006.

\bibitem[Bommasani et~al.(2022)Bommasani, Hudson, Adeli, Altman, Arora, von
  Arx, Bernstein, Bohg, Bosselut, Brunskill, Brynjolfsson, Buch, Card,
  Castellon, Chatterji, Chen, Creel, Davis, Demszky, Donahue, Doumbouya,
  Durmus, Ermon, Etchemendy, Ethayarajh, Fei-Fei, Finn, Gale, Gillespie, Goel,
  Goodman, Grossman, Guha, Hashimoto, Henderson, Hewitt, Ho, Hong, Hsu, Huang,
  Icard, Jain, Jurafsky, Kalluri, Karamcheti, Keeling, Khani, Khattab, Koh,
  Krass, Krishna, Kuditipudi, Kumar, Ladhak, Lee, Lee, Leskovec, Levent, Li,
  Li, Ma, Malik, Manning, Mirchandani, Mitchell, Munyikwa, Nair, Narayan,
  Narayanan, Newman, Nie, Niebles, Nilforoshan, Nyarko, Ogut, Orr,
  Papadimitriou, Park, Piech, Portelance, Potts, Raghunathan, Reich, Ren, Rong,
  Roohani, Ruiz, Ryan, Ré, Sadigh, Sagawa, Santhanam, Shih, Srinivasan,
  Tamkin, Taori, Thomas, Tramèr, Wang, Wang, Wu, Wu, Wu, Xie, Yasunaga, You,
  Zaharia, Zhang, Zhang, Zhang, Zhang, Zheng, Zhou, and
  Liang]{bommasani_opportunities_2022}
Rishi Bommasani, Drew~A. Hudson, Ehsan Adeli, Russ Altman, Simran Arora, Sydney
  von Arx, Michael~S. Bernstein, Jeannette Bohg, Antoine Bosselut, Emma
  Brunskill, Erik Brynjolfsson, Shyamal Buch, Dallas Card, Rodrigo Castellon,
  Niladri Chatterji, Annie Chen, Kathleen Creel, Jared~Quincy Davis, Dora
  Demszky, Chris Donahue, Moussa Doumbouya, Esin Durmus, Stefano Ermon, John
  Etchemendy, Kawin Ethayarajh, Li~Fei-Fei, Chelsea Finn, Trevor Gale, Lauren
  Gillespie, Karan Goel, Noah Goodman, Shelby Grossman, Neel Guha, Tatsunori
  Hashimoto, Peter Henderson, John Hewitt, Daniel~E. Ho, Jenny Hong, Kyle Hsu,
  Jing Huang, Thomas Icard, Saahil Jain, Dan Jurafsky, Pratyusha Kalluri,
  Siddharth Karamcheti, Geoff Keeling, Fereshte Khani, Omar Khattab, Pang~Wei
  Koh, Mark Krass, Ranjay Krishna, Rohith Kuditipudi, Ananya Kumar, Faisal
  Ladhak, Mina Lee, Tony Lee, Jure Leskovec, Isabelle Levent, Xiang~Lisa Li,
  Xuechen Li, Tengyu Ma, Ali Malik, Christopher~D. Manning, Suvir Mirchandani,
  Eric Mitchell, Zanele Munyikwa, Suraj Nair, Avanika Narayan, Deepak
  Narayanan, Ben Newman, Allen Nie, Juan~Carlos Niebles, Hamed Nilforoshan,
  Julian Nyarko, Giray Ogut, Laurel Orr, Isabel Papadimitriou, Joon~Sung Park,
  Chris Piech, Eva Portelance, Christopher Potts, Aditi Raghunathan, Rob Reich,
  Hongyu Ren, Frieda Rong, Yusuf Roohani, Camilo Ruiz, Jack Ryan, Christopher
  Ré, Dorsa Sadigh, Shiori Sagawa, Keshav Santhanam, Andy Shih, Krishnan
  Srinivasan, Alex Tamkin, Rohan Taori, Armin~W. Thomas, Florian Tramèr,
  Rose~E. Wang, William Wang, Bohan Wu, Jiajun Wu, Yuhuai Wu, Sang~Michael Xie,
  Michihiro Yasunaga, Jiaxuan You, Matei Zaharia, Michael Zhang, Tianyi Zhang,
  Xikun Zhang, Yuhui Zhang, Lucia Zheng, Kaitlyn Zhou, and Percy Liang.
\newblock On the {Opportunities} and {Risks} of {Foundation} {Models}.
\newblock \emph{arXiv:2108.07258}, 2022.

\bibitem[Bouvrie(2006)]{bouvrie_notes_2006}
Jake Bouvrie.
\newblock Notes on convolutional neural networks.
\newblock 2006.

\bibitem[Boyd \& Vandenberghe(2004)Boyd and Vandenberghe]{boyd_convex_2004}
Stephen Boyd and Lieven Vandenberghe.
\newblock \emph{Convex optimization}.
\newblock 2004.

\bibitem[Bulo et~al.(2018)Bulo, Porzi, and Kontschieder]{bulo_-place_2018}
Samuel~Rota Bulo, Lorenzo Porzi, and Peter Kontschieder.
\newblock In-place activated batchnorm for memory-optimized training of dnns.
\newblock In \emph{{CVPR}}, 2018.

\bibitem[Chen et~al.(2019)Chen, Wang, Pang, Cao, Xiong, Li, Sun, Feng, Liu, and
  Xu]{chen_mmdetection:_2019}
Kai Chen, Jiaqi Wang, Jiangmiao Pang, Yuhang Cao, Yu~Xiong, Xiaoxiao Li,
  Shuyang Sun, Wansen Feng, Ziwei Liu, and Jiarui Xu.
\newblock {MMDetection}: {Open} mmlab detection toolbox and benchmark.
\newblock \emph{arXiv preprint arXiv:1906.07155}, 2019.

\bibitem[Contributors(2018)]{mmcv}
MMCV Contributors.
\newblock {MMCV: OpenMMLab} computer vision foundation.
\newblock \url{https://github.com/open-mmlab/mmcv}, 2018.

\bibitem[Contributors(2022)]{mmengine2022}
MMEngine Contributors.
\newblock {MMEngine}: Openmmlab foundational library for training deep learning
  models.
\newblock 2022.

\bibitem[Dao et~al.(2022)Dao, Fu, Ermon, Rudra, and
  Ré]{dao_flashattention:_2022}
Tri Dao, Dan Fu, Stefano Ermon, Atri Rudra, and Christopher Ré.
\newblock Flashattention: {Fast} and memory-efficient exact attention with
  io-awareness.
\newblock In \emph{{NeurIPS}}, 2022.

\bibitem[Deng et~al.(2009)Deng, Dong, Socher, Li, Li, and
  Fei-Fei]{deng_imagenet:_2009}
Jia Deng, Wei Dong, Richard Socher, Li-Jia Li, Kai Li, and Li~Fei-Fei.
\newblock Imagenet: {A} large-scale hierarchical image database.
\newblock In \emph{{CVPR}}, 2009.

\bibitem[Ding et~al.(2021)Ding, Zhang, Ma, Han, Ding, and
  Sun]{ding_repvgg:_2021}
Xiaohan Ding, Xiangyu Zhang, Ningning Ma, Jungong Han, Guiguang Ding, and Jian
  Sun.
\newblock Repvgg: {Making} vgg-style convnets great again.
\newblock In \emph{{CVPR}}, 2021.

\bibitem[Donahue et~al.(2014)Donahue, Jia, Vinyals, Hoffman, Zhang, Tzeng, and
  Darrell]{donahue_decaf:_2014}
Jeff Donahue, Yangqing Jia, Oriol Vinyals, Judy Hoffman, Ning Zhang, Eric
  Tzeng, and Trevor Darrell.
\newblock Decaf: {A} deep convolutional activation feature for generic visual
  recognition.
\newblock In \emph{{ICML}}, 2014.

\bibitem[Gao et~al.(2019)Gao, Cheng, Zhao, Zhang, Yang, and
  Torr]{gao_res2net:_2019}
Shang-Hua Gao, Ming-Ming Cheng, Kai Zhao, Xin-Yu Zhang, Ming-Hsuan Yang, and
  Philip Torr.
\newblock Res2net: {A} new multi-scale backbone architecture.
\newblock \emph{TPAMI}, 2019.

\bibitem[Goodfellow et~al.(2015)Goodfellow, Shlens, and
  Szegedy]{goodfellow_explaining_2015}
Ian~J. Goodfellow, Jonathon Shlens, and Christian Szegedy.
\newblock Explaining and harnessing adversarial examples.
\newblock In \emph{{ICLR}}, 2015.

\bibitem[Gupta et~al.(2019)Gupta, Schwing, and Hoiem]{gupta_no-frills_2019}
Tanmay Gupta, Alexander Schwing, and Derek Hoiem.
\newblock No-frills human-object interaction detection: {Factorization}, layout
  encodings, and training techniques.
\newblock In \emph{{CVPR}}, 2019.

\bibitem[Han et~al.(2020)Han, Wang, Tian, Guo, Xu, and Xu]{han_ghostnet:_2020}
Kai Han, Yunhe Wang, Qi~Tian, Jianyuan Guo, Chunjing Xu, and Chang Xu.
\newblock Ghostnet: {More} features from cheap operations.
\newblock In \emph{{CVPR}}, 2020.

\bibitem[He et~al.(2017)He, Gkioxari, Dollár, and Girshick]{he_mask_2017}
K.~He, G.~Gkioxari, P.~Dollár, and R.~Girshick.
\newblock Mask {R}-{CNN}.
\newblock In \emph{{ICCV}}, 2017.

\bibitem[He et~al.(2016)He, Zhang, Ren, and Sun]{he_deep_2016}
Kaiming He, Xiangyu Zhang, Shaoqing Ren, and Jian Sun.
\newblock Deep residual learning for image recognition.
\newblock In \emph{{CVPR}}, 2016.

\bibitem[Huang et~al.(2023)Huang, Qin, Zhou, Zhu, Liu, and
  Shao]{huang_normalization_2023}
Lei Huang, Jie Qin, Yi~Zhou, Fan Zhu, Li~Liu, and Ling Shao.
\newblock Normalization techniques in training dnns: {Methodology}, analysis
  and application.
\newblock \emph{TPAMI}, 2023.

\bibitem[Ioffe(2017)]{ioffe_batch_2017}
Sergey Ioffe.
\newblock Batch renormalization: {Towards} reducing minibatch dependence in
  batch-normalized models.
\newblock In \emph{NeurIPS}, 2017.

\bibitem[Ioffe \& Szegedy(2015)Ioffe and Szegedy]{ioffe_batch_2015}
Sergey Ioffe and Christian Szegedy.
\newblock Batch {Normalization}: {Accelerating} {Deep} {Network} {Training} by
  {Reducing} {Internal} {Covariate} {Shift}.
\newblock In \emph{{ICML}}, 2015.

\bibitem[Jiang et~al.(2022)Jiang, Shu, Wang, and
  Long]{jiang2022transferability}
Junguang Jiang, Yang Shu, Jianmin Wang, and Mingsheng Long.
\newblock Transferability in deep learning: A survey.
\newblock \emph{arXiv preprint arXiv:2201.05867}, 2022.

\bibitem[Krause et~al.(2013)Krause, Stark, Deng, and Fei-Fei]{krause_3d_2013}
Jonathan Krause, Michael Stark, Jia Deng, and Li~Fei-Fei.
\newblock 3d {Object} {Representations} for {Fine}-{Grained} {Categorization}.
\newblock In \emph{{ICCV} {Workshop}}, 2013.

\bibitem[Kurakin et~al.(2016)Kurakin, Goodfellow, and
  Bengio]{kurakin2016adversarial}
Alexey Kurakin, Ian Goodfellow, and Samy Bengio.
\newblock Adversarial machine learning at scale.
\newblock \emph{arXiv preprint arXiv:1611.01236}, 2016.

\bibitem[Kuznetsova et~al.(2018)Kuznetsova, Rom, Alldrin, Uijlings, Krasin,
  Pont-Tuset, Kamali, Popov, Malloci, and Duerig]{kuznetsova_open_2018}
Alina Kuznetsova, Hassan Rom, Neil Alldrin, Jasper Uijlings, Ivan Krasin, Jordi
  Pont-Tuset, Shahab Kamali, Stefan Popov, Matteo Malloci, and Tom Duerig.
\newblock The open images dataset v4: {Unified} image classification, object
  detection, and visual relationship detection at scale.
\newblock \emph{arXiv preprint arXiv:1811.00982}, 2018.

\bibitem[LeCun et~al.(1998)LeCun, Bottou, Bengio, and
  Haffner]{lecun_gradient-based_1998}
Yann LeCun, Léon Bottou, Yoshua Bengio, and Patrick Haffner.
\newblock Gradient-based learning applied to document recognition.
\newblock \emph{Proc. IEEE}, 1998.

\bibitem[Lin et~al.(2014)Lin, Maire, Belongie, Hays, Perona, Ramanan, Dollár,
  and Zitnick]{lin_microsoft_2014}
Tsung-Yi Lin, Michael Maire, Serge Belongie, James Hays, Pietro Perona, Deva
  Ramanan, Piotr Dollár, and C.~Lawrence Zitnick.
\newblock Microsoft coco: {Common} objects in context.
\newblock In \emph{{ECCV}}, 2014.

\bibitem[Lin et~al.(2017)Lin, Goyal, Girshick, He, and Dollár]{lin_focal_2017}
Tsung-Yi Lin, Priya Goyal, Ross Girshick, Kaiming He, and Piotr Dollár.
\newblock Focal loss for dense object detection.
\newblock In \emph{{ICCV}}, 2017.

\bibitem[Maas et~al.(2013)Maas, Hannun, and Ng]{maas_rectifier_2013}
Andrew~L. Maas, Awni~Y. Hannun, and Andrew~Y. Ng.
\newblock Rectifier nonlinearities improve neural network acoustic models.
\newblock In \emph{{ICML}}, 2013.

\bibitem[Madry et~al.(2017)Madry, Makelov, Schmidt, Tsipras, and
  Vladu]{madry2017towards}
Aleksander Madry, Aleksandar Makelov, Ludwig Schmidt, Dimitris Tsipras, and
  Adrian Vladu.
\newblock Towards deep learning models resistant to adversarial attacks.
\newblock \emph{arXiv preprint arXiv:1706.06083}, 2017.

\bibitem[Mahajan et~al.(2018)Mahajan, Girshick, Ramanathan, He, Paluri, Li,
  Bharambe, and van~der Maaten]{mahajan_exploring_2018}
Dhruv Mahajan, Ross Girshick, Vignesh Ramanathan, Kaiming He, Manohar Paluri,
  Yixuan Li, Ashwin Bharambe, and Laurens van~der Maaten.
\newblock Exploring the limits of weakly supervised pretraining.
\newblock In \emph{{ECCV}}, 2018.

\bibitem[Maji et~al.(2013)Maji, Rahtu, Kannala, Blaschko, and
  Vedaldi]{maji2013fine}
Subhransu Maji, Esa Rahtu, Juho Kannala, Matthew Blaschko, and Andrea Vedaldi.
\newblock Fine-grained visual classification of aircraft.
\newblock \emph{arXiv preprint arXiv:1306.5151}, 2013.

\bibitem[Markuš(2018)]{markus_fusing_nodate}
Nenad Markuš.
\newblock Fusing batch normalization and convolution in runtime.
\newblock \emph{https://nenadmarkus.com/p/fusing-batchnorm-and-conv/}, 2018.

\bibitem[Mehta et~al.(2020)Mehta, Sotnychenko, Mueller, Xu, Elgharib, Fua,
  Seidel, Rhodin, Pons-Moll, and Theobalt]{mehta2020xnect}
Dushyant Mehta, Oleksandr Sotnychenko, Franziska Mueller, Weipeng Xu, Mohamed
  Elgharib, Pascal Fua, Hans-Peter Seidel, Helge Rhodin, Gerard Pons-Moll, and
  Christian Theobalt.
\newblock Xnect: Real-time multi-person 3d motion capture with a single rgb
  camera.
\newblock \emph{TOG}, 2020.

\bibitem[Micikevicius et~al.(2018)Micikevicius, Narang, Alben, Diamos, Elsen,
  Garcia, Ginsburg, Houston, Kuchaiev, Venkatesh, and
  Wu]{micikevicius_mixed_2018}
Paulius Micikevicius, Sharan Narang, Jonah Alben, Gregory Diamos, Erich Elsen,
  David Garcia, Boris Ginsburg, Michael Houston, Oleksii Kuchaiev, Ganesh
  Venkatesh, and Hao Wu.
\newblock Mixed {Precision} {Training}.
\newblock In \emph{{ICLR}}, 2018.

\bibitem[Nair \& Hinton(2010)Nair and Hinton]{nair_rectified_2010}
Vinod Nair and Geoffrey~E. Hinton.
\newblock Rectified linear units improve restricted boltzmann machines.
\newblock In \emph{{ICML}}, 2010.

\bibitem[Paszke et~al.(2019)Paszke, Gross, Massa, Lerer, Bradbury, Chanan,
  Killeen, Lin, Gimelshein, Antiga, Desmaison, Kopf, Yang, DeVito, Raison,
  Tejani, Chilamkurthy, Steiner, Fang, Bai, and Chintala]{paszke_pytorch:_2019}
Adam Paszke, Sam Gross, Francisco Massa, Adam Lerer, James Bradbury, Gregory
  Chanan, Trevor Killeen, Zeming Lin, Natalia Gimelshein, Luca Antiga, Alban
  Desmaison, Andreas Kopf, Edward Yang, Zachary DeVito, Martin Raison, Alykhan
  Tejani, Sasank Chilamkurthy, Benoit Steiner, Lu~Fang, Junjie Bai, and Soumith
  Chintala.
\newblock {PyTorch}: {An} {Imperative} {Style}, {High}-{Performance} {Deep}
  {Learning} {Library}.
\newblock In \emph{{NeurIPS}}, 2019.

\bibitem[Rabiner \& Gold(1975)Rabiner and Gold]{rabiner_theory_1975}
Lawrence~R. Rabiner and Bernard Gold.
\newblock Theory and application of digital signal processing.
\newblock \emph{Englewood Cliffs: Prentice-Hall}, 1975.

\bibitem[Radosavovic et~al.(2020)Radosavovic, Kosaraju, Girshick, He, and
  Doll{\'a}r]{radosavovic2020designing}
Ilija Radosavovic, Raj~Prateek Kosaraju, Ross Girshick, Kaiming He, and Piotr
  Doll{\'a}r.
\newblock Designing network design spaces.
\newblock In \emph{{CVPR}}, 2020.

\bibitem[Reed et~al.(2022)Reed, DeVito, He, Ussery, and
  Ansel]{reed_torch._2022}
James Reed, Zachary DeVito, Horace He, Ansley Ussery, and Jason Ansel.
\newblock torch. fx: {Practical} {Program} {Capture} and {Transformation} for
  {Deep} {Learning} in {Python}.
\newblock In \emph{{MLSys}}, 2022.

\bibitem[Ren et~al.(2015)Ren, He, Girshick, and Sun]{ren_faster_2015}
Shaoqing Ren, Kaiming He, Ross Girshick, and Jian Sun.
\newblock Faster {R}-{CNN}: {Towards} {Real}-{Time} {Object} {Detection} with
  {Region} {Proposal} {Networks}.
\newblock In \emph{{NeurIPS}}, 2015.

\bibitem[Ronneberger et~al.(2015)Ronneberger, Fischer, and
  Brox]{ronneberger2015u}
Olaf Ronneberger, Philipp Fischer, and Thomas Brox.
\newblock U-net: Convolutional networks for biomedical image segmentation.
\newblock In \emph{{MICCAI}}, 2015.

\bibitem[Sun et~al.(2019)Sun, Xiao, Liu, and Wang]{sun2019deep}
Ke~Sun, Bin Xiao, Dong Liu, and Jingdong Wang.
\newblock Deep high-resolution representation learning for human pose
  estimation.
\newblock In \emph{{CVPR}}, 2019.

\bibitem[Szegedy et~al.(2013)Szegedy, Zaremba, Sutskever, Bruna, Erhan,
  Goodfellow, and Fergus]{szegedy2013intriguing}
Christian Szegedy, Wojciech Zaremba, Ilya Sutskever, Joan Bruna, Dumitru Erhan,
  Ian Goodfellow, and Rob Fergus.
\newblock Intriguing properties of neural networks.
\newblock \emph{arXiv preprint arXiv:1312.6199}, 2013.

\bibitem[Vanholder(2016)]{vanholder_efficient_2016}
Han Vanholder.
\newblock Efficient inference with tensorrt.
\newblock In \emph{{GPU} {Technology} {Conference}}, 2016.

\bibitem[Wah et~al.(2011)Wah, Branson, Welinder, Perona, and
  Belongie]{wah2011caltech}
Catherine Wah, Steve Branson, Peter Welinder, Pietro Perona, and Serge
  Belongie.
\newblock The caltech-ucsd birds-200-2011 dataset.
\newblock 2011.

\bibitem[Wang et~al.(2021)Wang, Shelhamer, Liu, Olshausen, and
  Darrell]{wang_tent:_2021}
Dequan Wang, Evan Shelhamer, Shaoteng Liu, Bruno Olshausen, and Trevor Darrell.
\newblock Tent: {Fully} {Test}-{Time} {Adaptation} by {Entropy} {Minimization}.
\newblock In \emph{{ICLR}}, 2021.

\bibitem[Wang et~al.(2022)Wang, Fink, Van~Gool, and Dai]{wang_continual_2022}
Qin Wang, Olga Fink, Luc Van~Gool, and Dengxin Dai.
\newblock Continual {Test}-{Time} {Domain} {Adaptation}.
\newblock In \emph{{CVPR}}, 2022.

\bibitem[Wang et~al.(2019)Wang, Jin, Long, Wang, and
  Jordan]{wang_transferable_2019}
Ximei Wang, Ying Jin, Mingsheng Long, Jianmin Wang, and Michael~I. Jordan.
\newblock Transferable {Normalization}: {Towards} {Improving} {Transferability}
  of {Deep} {Neural} {Networks}.
\newblock In \emph{{NeurIPS}}, 2019.

\bibitem[Wu(2023)]{wu_pytorch_2023}
Peng Wu.
\newblock {PyTorch} 2.0: {The} {Journey} to {Bringing} {Compiler}
  {Technologies} to the {Core} of {PyTorch}.
\newblock In \emph{{CGO}}, 2023.

\bibitem[Wu \& He(2018)Wu and He]{wu_group_2018}
Yuxin Wu and Kaiming He.
\newblock Group {Normalization}.
\newblock In \emph{{ECCV}}, 2018.

\bibitem[Wu et~al.(2019)Wu, Kirillov, Massa, Lo, and
  Girshick]{wu2019detectron2}
Yuxin Wu, Alexander Kirillov, Francisco Massa, Wan-Yen Lo, and Ross Girshick.
\newblock Detectron2.
\newblock \url{https://github.com/facebookresearch/detectron2}, 2019.

\bibitem[Xie et~al.(2017)Xie, Girshick, Doll{\'a}r, Tu, and
  He]{xie2017aggregated}
Saining Xie, Ross Girshick, Piotr Doll{\'a}r, Zhuowen Tu, and Kaiming He.
\newblock Aggregated residual transformations for deep neural networks.
\newblock In \emph{{CVPR}}, 2017.

\bibitem[Yu et~al.(2018)Yu, Wang, Shelhamer, and Darrell]{yu2018deep}
Fisher Yu, Dequan Wang, Evan Shelhamer, and Trevor Darrell.
\newblock Deep layer aggregation.
\newblock In \emph{{CVPR}}, 2018.

\bibitem[Zeiler et~al.(2010)Zeiler, Krishnan, Taylor, and
  Fergus]{zeiler_deconvolutional_2010}
Matthew~D. Zeiler, Dilip Krishnan, Graham~W. Taylor, and Rob Fergus.
\newblock Deconvolutional networks.
\newblock In \emph{{CVPR}}, 2010.

\bibitem[Zhang et~al.(2022)Zhang, Wu, Zhang, Zhu, Lin, Zhang, Sun, He, Mueller,
  Manmatha, et~al.]{zhang2022resnest}
Hang Zhang, Chongruo Wu, Zhongyue Zhang, Yi~Zhu, Haibin Lin, Zhi Zhang, Yue
  Sun, Tong He, Jonas Mueller, R~Manmatha, et~al.
\newblock Resnest: Split-attention networks.
\newblock In \emph{{CVPR}}, 2022.

\bibitem[Zhou et~al.(2018)Zhou, Lapedriza, Khosla, Oliva, and
  Torralba]{zhou_places:_2018}
B.~Zhou, A.~Lapedriza, A.~Khosla, A.~Oliva, and A.~Torralba.
\newblock Places: {A} 10 {Million} {Image} {Database} for {Scene}
  {Recognition}.
\newblock \emph{TPAMI}, 2018.

\end{thebibliography}
